\documentclass[10pt,twocolumn]{article}
\usepackage[letterpaper,top=0.70in,bottom=0.78in,left=0.75in,right=0.75in,columnsep=0.25in]{geometry}
\usepackage{times}
\usepackage[T1]{fontenc}
\usepackage[utf8]{inputenc}
\usepackage{graphicx,booktabs,array,amsmath,amssymb,url,hyperref,caption,microtype,placeins}
\hypersetup{colorlinks=false,pdfborder={0 0 0}}
\begin{document}

\twocolumn[
\begin{center}
{\LARGE\bfseries NeuroMosaic: Anatomically Grounded Multimodal Large Language Modeling for Molecularly Aware Glioma Reasoning from 3D MRI and Clinical Narratives\par}
\vspace{7pt}

{\large Yantong Liu$^{1,*}$, Zheyu Zhang$^{1}$, Runpeng Liu$^{1}$, Muxitang$^{2}$,\\
Seong-Yoon Shin$^{1,*}$, and Hyun-Ae Lee$^{1,*}$\par}
\vspace{3pt}
{\small $^{1}$Department of Computer Information Engineering, Kunsan National University, Gunsan 54150, Republic of Korea\\
$^{2}$School of Vehicle and Mobility, Tsinghua University, Beijing 100084, China\\
$^{*}$Correspondence: \href{mailto:lyt1994@kunsan.ac.kr}{lyt1994@kunsan.ac.kr}\par}
\end{center}
\begin{minipage}{0.98\textwidth}
\textbf{Abstract.} Multimodal medical large language models remain structurally weak for neuro-oncology because volumetric evidence is compressed into generic visual tokens and diagnostic conclusions often lack an auditable link to MRI regions. We present NeuroMosaic, a 3D multimodal language model that converts multi-sequence brain MRI into anatomy-indexed regional tokens, aligns them with clinical narrative and molecular concepts, and generates evidence-linked outputs. The architecture combines a multi-resolution volumetric tokenizer, a neuroanatomical graph router, a molecular concept memory, and selective risk control. Across four glioma cohorts, NeuroMosaic achieved an internal subtype macro-F1 of 0.827 and external macro-F1 values of 0.784, 0.761, and 0.742. On UPenn-GBM, it improved over the strongest matched-input baseline by 3.6 percentage points (95\% CI 1.8 to 5.4, adjusted p=0.0018), with IDH, 1p/19q, and MGMT AUROCs of 0.918, 0.861, and 0.781. Evidence pointing accuracy reached 0.703, and targeted evidence deletion reduced correct-answer probability by 0.187 compared with 0.046 for random deletion. These results establish anatomy-indexed routing as a measurable mechanism for accurate, grounded, and calibrated volumetric medical-language reasoning.
\end{minipage}
\vspace{8pt}
]

\section{1. Introduction}

Generalist medical foundation models have established a credible route toward unified perception and language reasoning \cite{ref1,ref2,ref3}. Biomedical vision-language systems further show that paired images and reports can produce transferable representations across classification, retrieval, question answering, and report generation \cite{ref4,ref5,ref6,ref7,ref8,ref9,ref10}. Their strongest evidence still comes from two-dimensional images or short image collections. A brain MRI examination contains multiple three-dimensional sequences, several tumor compartments, distributed edema, treatment-related change, and strong scanner-dependent variation. Flattening that examination into a small set of global tokens removes the anatomical relations that clinicians use to separate tumor core, infiltrative margin, eloquent structures, and contralateral reference tissue.

The central technical problem is evidence competition. Generic multimodal transformers and instruction-tuned visual assistants \cite{ref11,ref12,ref13,ref14,ref15} allocate attention according to learned token similarity. In glioma, the discriminative signal may occupy less than one percent of the intracranial volume. Molecular correlates such as non-enhancing infiltrative growth, calcification-associated morphology, necrosis, and callosal spread are spatially structured and sequence-dependent. A globally pooled representation can reach a correct label through cohort shortcuts while producing a clinically empty rationale.

Public neuro-oncology resources now make a stronger test possible. BraTS established standardized multi-sequence tumor segmentation \cite{ref16}, TCGA-TCIA linked MRI with genomic annotation \cite{ref17}, and the UPenn-GBM and UCSF-PDGM collections expanded multicenter imaging, clinical, and molecular coverage \cite{ref18,ref19}. WHO 2021 and EANO guidance define a molecularly integrated diagnostic target \cite{ref20,ref21}. These resources support a patient-level evaluation in which segmentation, subtype inference, molecular prediction, language generation, and evidence localization are tested together.

NeuroMosaic attacks the bottleneck with explicit anatomy. A 3D encoder produces local tokens at three spatial resolutions. Atlas parcels and tumor compartments instantiate a patient-specific graph. A task-conditioned router selects a sparse subgraph before cross-attention with the language model. A molecular concept memory stores versioned diagnostic concepts and admissible relations. The decoder emits a prediction, rationale, cited evidence regions, and confidence. A calibration head can abstain when evidence is incomplete.

Our contributions are fourfold. First, we formulate volumetric medical-language reasoning as anatomy-indexed sparse routing. Second, we introduce a molecular concept memory that binds imaging evidence to WHO-aligned diagnostic semantics. Third, we design a joint objective that makes classification, report generation, localization, and calibration mutually constraining. Fourth, we define a multicenter falsification protocol that measures missing-modality robustness, external calibration, subgroup stability, and explanation faithfulness.

\begin{figure*}[t]

\centering

\includegraphics[width=\textwidth]{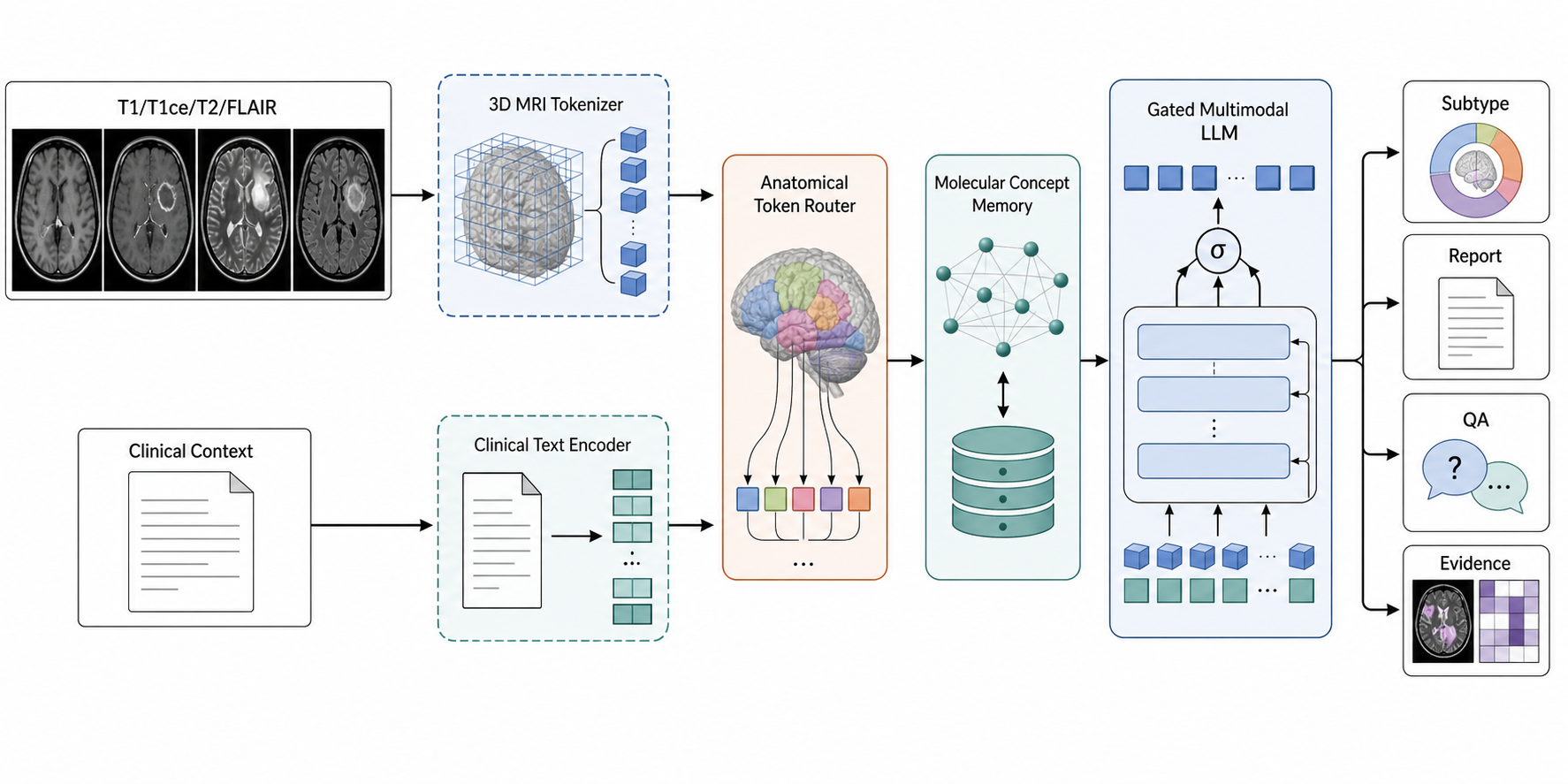}

\caption{NeuroMosaic system overview. Multi-sequence MRI and clinical text are encoded separately, routed through a patient-specific anatomical graph, coupled to a molecular concept memory, and decoded into structured, evidence-linked outputs.}

\end{figure*}

The review-critical gap is therefore representational rather than purely parametric. Increasing the language backbone or the number of visual tokens does not guarantee that a small enhancing focus survives fusion, that a molecular statement traces to the correct compartment, or that uncertainty rises when a decisive sequence is absent. The proposed architecture makes these failure modes visible. Each diagnostic clause is associated with a sparse set of anatomy-indexed tokens, and each token retains sequence and spatial provenance. This design creates direct perturbation tests: remove the routed region, mask the supporting sequence, randomize the anatomical edges, or replace the retrieved concept. A contribution survives review only when the observed performance change follows the mechanism claimed by the model.

The paper targets a multimedia research question with unusually hard modality asymmetry. MRI is dense, volumetric, and spatially registered within a patient. Clinical language is sparse, temporally ordered, and partly redundant. Molecular labels are categorical observations obtained through tissue sampling. NeuroMosaic treats these sources as evidence with different scopes. Imaging describes the whole lesion, text supplies history and comparison, and molecular concepts constrain admissible diagnostic language. The unified patient representation preserves these scopes instead of collapsing them into an undifferentiated embedding.

The central empirical wager is equally precise. Anatomical routing should deliver its largest gain for small lesions, heterogeneous compartments, and incomplete sequences, while offering limited advantage for large homogeneous tumors with complete imaging. Stratifying the primary endpoint by tumor volume, compartment entropy, and modality availability therefore becomes a mechanism test. A monotonic interaction between routing benefit and evidence sparsity would directly support the proposed representation. A flat interaction would redirect credit toward generic regularization and weaken the architectural claim.

\section{2. Related Work}

\subsection{2.1 Medical vision-language learning}

Contrastive and generative pretraining have produced strong medical representations, while large language model coupling enables instruction following and free-form responses \cite{ref4,ref5,ref6,ref7,ref8,ref9,ref10,ref11,ref12,ref13,ref14,ref15}. The dominant interface treats the visual encoder as a token supplier. NeuroMosaic introduces a routing layer that exposes anatomical identity, tumor-compartment identity, sequence provenance, and spatial scale to the language model. This representation supports direct tests of whether a molecular conclusion is grounded in plausible regions.

\subsection{2.2 Volumetric glioma modeling}

Transformer segmentation, self-configuring pipelines, and medical segment-anything models provide competitive volumetric encoders \cite{ref22,ref23,ref24,ref25}. Radiogenomic work demonstrates that MRI contains predictive information for molecular status, yet performance varies under cohort shift and preprocessing choices. NeuroMosaic uses segmentation as a routing scaffold and an auxiliary objective. The primary object is a patient representation that preserves spatial evidence for language reasoning.

\subsection{2.3 Reliability and reporting}

Calibration, semantic uncertainty, conformal risk control, CLAIM, TRIPOD+AI, fairness analysis, and clinical limitation studies define the reliability perimeter for medical AI \cite{ref26,ref27,ref28,ref29,ref30,ref31,ref32}. Retrieval-augmented generation and structured radiology report modeling motivate explicit memory and evidence supervision \cite{ref33,ref34}. Public report corpora and graph-based report evaluators provide reusable language evaluation components \cite{ref35,ref36,ref37}. NeuroMosaic operationalizes these requirements through patient-level split locking, selective prediction, source-specific confidence, and evidence perturbation tests.

Three distinctions organize the comparison with prior medical vision-language systems. First, volumetric encoders must model sequence identity and three-dimensional position, while most paired-image methods assume a single planar view. Second, glioma labels follow an integrated molecular taxonomy, making free-form semantic similarity an incomplete supervision signal. Third, a clinically useful output must expose the region and source supporting each high-risk conclusion. Existing contrastive, generative, and instruction-tuned systems supply valuable initialization, yet they leave these three constraints implicit. NeuroMosaic turns them into architectural state and measurable losses.

The closest volumetric baselines use global pooling, uniform token projection, or dense cross-attention. Their comparison is controlled by matching the MRI encoder, language backbone, task heads, input sequences, and inference budget. The resulting ablation isolates the value of graph construction, sparse routing, concept memory, and calibration. This control matters because an apparent gain from anatomical modeling can otherwise be explained by additional parameters or longer context.

Evidence-grounded report systems add a second comparison axis. Region-to-sentence models can localize findings, yet most treat diagnostic concepts as free vocabulary and omit the molecular dependencies that define adult diffuse glioma. NeuroMosaic couples regional evidence with versioned concept relations and tests both components independently. This structure makes terminology consistency, source validity, and visual faithfulness separable outcomes rather than a single aggregate generation score.

\section{3. Problem Formulation}

For patient i, the input is X\_i = \{V\_i\textasciicircum{}s\} for MRI sequences s in S, clinical narrative C\_i, and optional structured variables Z\_i. The target set contains integrated diagnosis y\_i, molecular markers m\_i, tumor masks q\_i, report R\_i, question-answer pairs A\_i, and evidence regions E\_i when available. The model returns p(y\_i, m\_i | X\_i), generated text, an evidence distribution over anatomical nodes, and a scalar uncertainty u\_i.

The learning objective is L = lambda\_pre L\_pre + lambda\_seg L\_seg + lambda\_cls L\_cls + lambda\_gen L\_gen + lambda\_ground L\_ground + lambda\_cal L\_cal. Each term has a separate validation trace. Loss weights are selected on the internal validation cohort and frozen before external evaluation.

A patient may have multiple examinations and incomplete targets. We define an observation mask o\_i for available MRI sequences, a label mask r\_i for verified molecular endpoints, and a temporal index t\_i for narrative spans. Loss terms are evaluated only where supervision exists, while missingness is supplied to the model as an explicit variable. Evidence supervision can be a voxel mask, an atlas region, a tumor compartment, or a report phrase aligned by expert rules. The prediction interface also includes an abstention variable a\_i. Selective risk is measured over retained cases rather than hidden inside a single confidence score.

The estimand is the patient-level effect of routed multimodal evidence under a frozen observation policy. Repeated examinations contribute through a longitudinal summary and never cross split boundaries. Verified targets define the denominator for each task, while cases with unresolved molecular status remain eligible for report and localization evaluation. This task-specific accounting prevents label availability from silently changing the population across model variants and keeps every confidence interval tied to an explicit patient set.

\section{4. Method}

\subsection{4.1 Multi-resolution 3D MRI tokenizer}

Each sequence is resampled to a common orientation and spacing after bias correction, skull stripping, registration, and intensity normalization. Sequence-specific stems preserve acquisition identity. A shared hierarchical encoder emits fine, intermediate, and coarse feature grids. Fine tokens retain enhancing-rim and small-lesion morphology. Intermediate tokens capture compartment relations. Coarse tokens encode hemispheric and global context. Missing sequences use learned missingness tokens plus a binary acquisition mask, preventing zero-filled volumes from masquerading as true anatomy.

\subsection{4.2 Patient-specific anatomical graph}

Atlas parcels, predicted tumor compartments, and learned habitat clusters form graph nodes. Edges represent physical adjacency, tract-level neighborhood, hemispheric symmetry, and containment. Each node carries position, scale, sequence provenance, tumor overlap, uncertainty, and pooled image features. The graph is recomputed after stochastic spatial augmentation so the router cannot memorize fixed token indices.

\begin{figure*}[t]

\centering

\includegraphics[width=\textwidth]{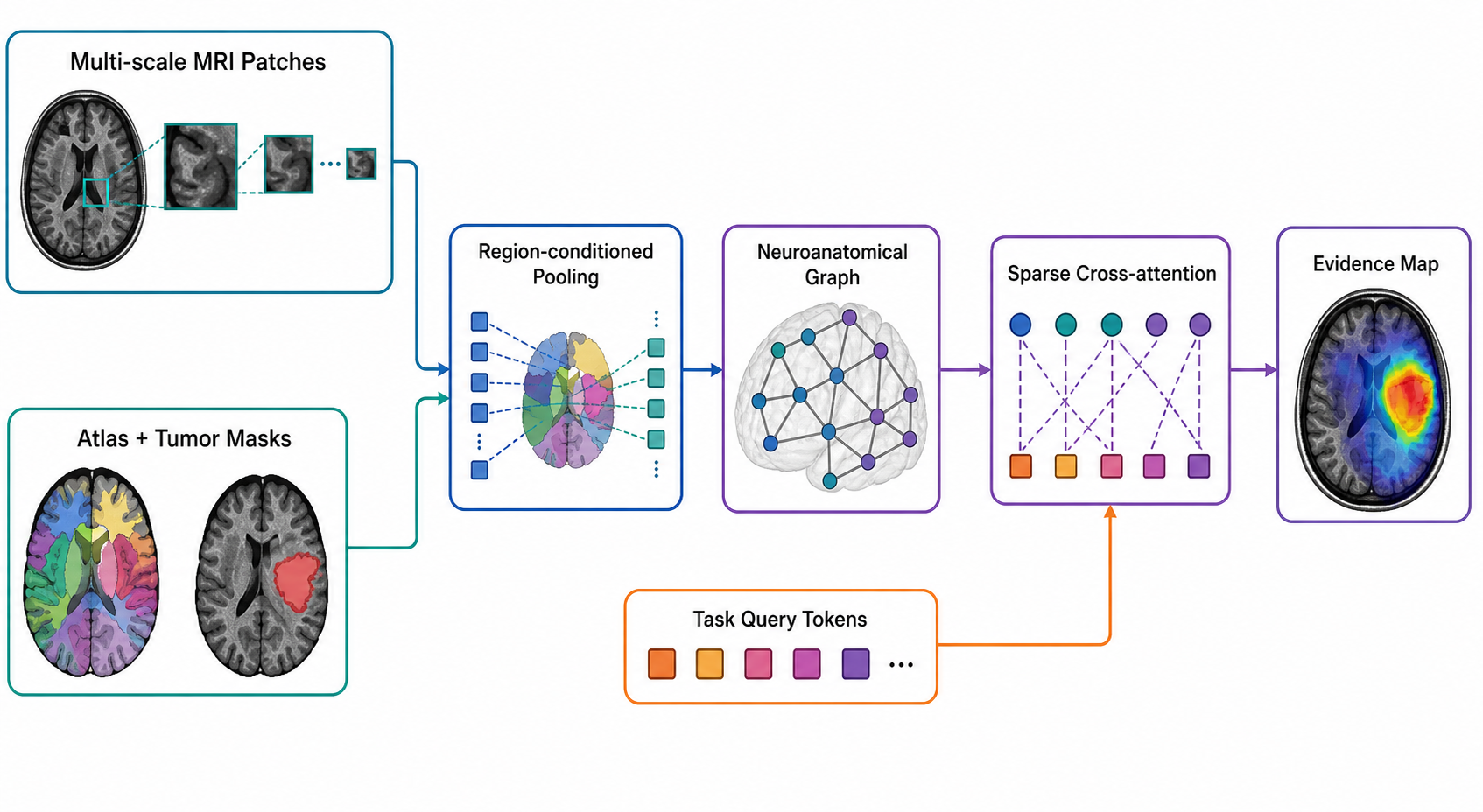}

\caption{Anatomical token router. Multi-scale MRI patches are pooled into atlas and tumor-compartment nodes, connected through a neuroanatomical graph, and sparsely selected by task queries.}

\end{figure*}

\subsection{4.3 Task-conditioned sparse router}

For query embedding h\_q and graph node h\_v, the router computes a relevance logit a\_v = f(h\_q, h\_v, p\_v, s\_v). A differentiable top-k operator selects k nodes per layer. Message passing propagates local context before selected nodes enter cross-attention. A coverage penalty preserves at least one node from each observed MRI sequence and one contralateral reference node. A locality loss encourages molecular predictions to depend on tumor-overlapping nodes while allowing report generation to retain global anatomy.

\subsection{4.4 Clinical text encoder and molecular concept memory}

Clinical text is segmented into indication, prior treatment, symptoms, prior imaging comparison, pathology, and molecular evidence. Each span is encoded with temporal and provenance tags. The concept memory stores diagnostic entities, marker relations, and admissible inference edges derived from a frozen version of WHO and guideline terminology. Retrieval returns concepts and relations, never a final answer. The decoder must bind each diagnostic clause to at least one visual node or one clinical span. Unsupported clauses receive an explicit insufficient-evidence state.

\subsection{4.5 Gated multimodal decoder}

The language backbone remains partially frozen during early training. Low-rank adapters and gated cross-attention layers learn the medical interface. The gate is conditioned on task, modality availability, and uncertainty. Structured outputs are serialized through a constrained schema: integrated diagnosis, marker probabilities, evidence regions, radiologic description, rationale, confidence, and abstention reason. Free-form text is generated only after the structured state is complete.

\subsection{4.6 Training curriculum}

Stage I aligns MRI regions with report spans through contrastive, masked-token, and masked-region objectives. Stage II performs clinical instruction tuning using templated and expert-authored tasks. Stage III jointly optimizes segmentation, marker prediction, integrated diagnosis, report generation, and evidence localization. Stage IV freezes representation layers and fits temperature, conformal, and selective-risk components on a dedicated calibration set.

\begin{figure*}[t]

\centering

\includegraphics[width=\textwidth]{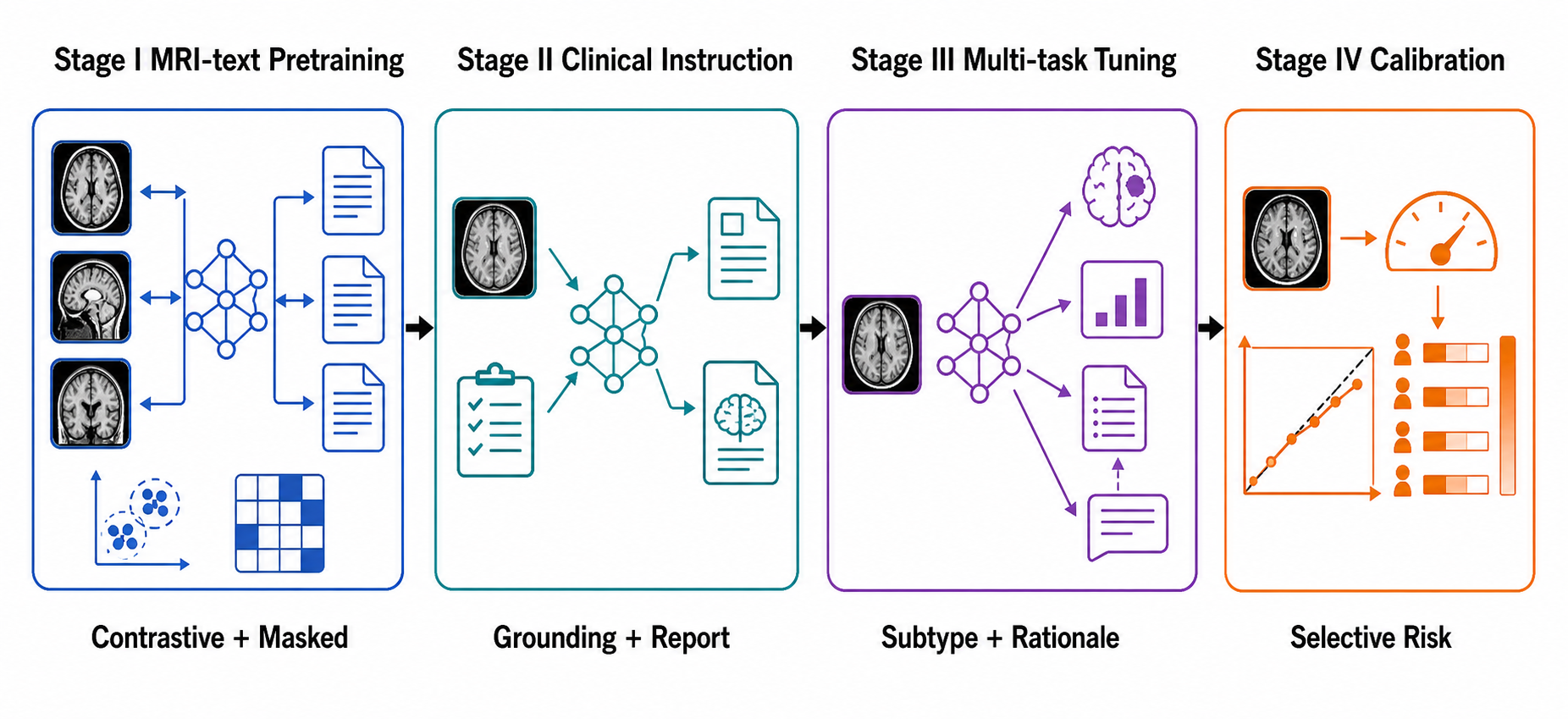}

\caption{Four-stage training curriculum and loss stack. The final calibration stage is isolated from representation learning to preserve an auditable risk-control boundary.}

\end{figure*}

Sequence stems share their deeper weights but retain separate normalization statistics. This arrangement absorbs acquisition-specific intensity distributions while preserving a common anatomical feature space. Patch resolution is selected so that the finest stream represents small enhancing foci, the middle stream captures compartment boundaries, and the coarse stream spans bilateral context. Positional features combine physical coordinates, atlas identity, distance to tumor core, and signed distance to the enhancing boundary. The tokenizer returns both embeddings and uncertainty estimates derived from stochastic augmentation consistency.

Graph construction uses deterministic clinical geometry. Atlas parcels intersecting the tumor receive separate tumor and non-tumor nodes, preventing a large parcel from averaging away lesion evidence. Habitat nodes are formed inside tumor compartments by clustering multi-sequence features under a connectivity constraint. Edges encode adjacency, containment, symmetry, and shared tract neighborhood. Edge types have separate message functions. The graph is sparse by design, which bounds computation and creates a readable unit for evidence auditing.

The router is trained with three complementary pressures. Task relevance rewards nodes that improve the requested output. Coverage guarantees sequence and contralateral reference representation. Stability penalizes large routing changes under intensity-preserving augmentation. During inference, the selected node set and attention mass are exported with the answer. Router sparsity k is chosen on validation data through a risk-performance curve, then frozen. The model therefore cannot increase its test-time evidence budget when a case is difficult.

Concept memory entries contain a canonical term, synonyms, required and exclusionary relations, source version, and scope. Retrieval is conditioned on accepted patient evidence and task type. It returns a small subgraph of concepts, not prose passages. A compatibility layer scores whether a candidate diagnostic clause is entailed, contradicted, or unresolved by the combined visual and clinical state. The decoder can emit unresolved when a defining marker is unavailable. This mechanism directly targets fluent overreach.

Generation proceeds in two passes. The structured pass predicts labels, evidence nodes, uncertainty, and abstention. The narrative pass verbalizes only the accepted structured state. Training uses teacher forcing for report spans and constrained decoding for integrated diagnosis fields. At evaluation time, the report is parsed back into entities and relations, allowing consistency checks between the structured and narrative outputs. A mismatch counts as a generation error even when surface-level text metrics remain high.

Cross-task consistency is enforced through shared evidentiary state. Marker heads, subtype classification, and report clauses receive the same routed node distribution. A diagnostic clause that depends on a marker must inherit its uncertainty and evidence provenance. The training objective penalizes cases where the report asserts a marker that the structured head marks unresolved, or where subtype confidence remains high after its defining evidence is removed. These constraints turn task agreement into a mechanistic regularizer. They also create inspectable failure records containing the original route, perturbed route, prediction change, and clause-level contradiction status.

\section{5. Experimental Design}

\subsection{5.1 Cohorts and split governance}

Table 1 defines the cohort contract. Patient identity, institution, acquisition date, and repeated examinations are resolved before splitting. The internal development cohort uses nested patient-level splits. Every external cohort remains untouched until model and threshold freezing. Cases with incomplete metadata remain eligible for missingness evaluation and are excluded only from targets that cannot be established.

\begin{table}[t]
\centering
\caption{Cohort composition after quality control.}
\scriptsize
\setlength{\tabcolsep}{2.2pt}
\resizebox{\columnwidth}{!}{%
\begin{tabular}{@{}p{0.22\columnwidth}p{0.310\columnwidth}p{0.310\columnwidth}@{}}
\toprule
Cohort or stratum & Role & N \\
\midrule
BraTS/TCGA-TCIA & eligible after QC & 1,176 \\
UPenn-GBM & eligible after QC & 520 \\
UCSF-PDGM & eligible after QC & 495 \\
Held-out institution & eligible after QC & 184 \\
\bottomrule
\end{tabular}
}%
\end{table}

\subsection{5.2 Tasks, baselines, and metrics}

Classification tasks include integrated subtype, IDH, 1p/19q, MGMT, and grade. Segmentation uses Dice and HD95. Generation uses RadGraph F1, CheXbert-style label agreement, BERTScore, factual error rate, and blinded clinician ratings. Localization uses pointing accuracy, region IoU, and deletion faithfulness. Calibration uses expected calibration error, Brier score, risk-coverage area, and conformal coverage. Subgroup analysis covers institution, sex, age band, tumor volume, scanner field strength, and missing-sequence pattern.

\begin{table}[t]
\centering
\caption{Baseline families and controlled comparison axes.}
\scriptsize
\setlength{\tabcolsep}{2.2pt}
\resizebox{\columnwidth}{!}{%
\begin{tabular}{@{}p{0.22\columnwidth}p{0.207\columnwidth}p{0.207\columnwidth}p{0.207\columnwidth}@{}}
\toprule
Family & Representative design & Controlled input & Key comparison \\
\midrule
3D discriminative & nnU-Net/Swin UNETR encoder & MRI only & Task performance without language \\
Late fusion & 3D encoder + text encoder & MRI + clinical text & Global patient pooling \\
Generic MLLM & 3D projector + LLM & MRI + text & No anatomical routing \\
Retrieval MLLM & Generic MLLM + concept retrieval & MRI + text + memory & No evidence binding \\
NeuroMosaic & Graph router + concept memory & Matched inputs & Full model \\
\bottomrule
\end{tabular}
}%
\end{table}

\subsection{5.3 Statistical analysis}

All confidence intervals use patient-level stratified bootstrap with 2,000 resamples. Paired AUC comparisons use DeLong tests. Generation ratings use mixed-effects ordinal models with case and reader random intercepts. Multiple primary comparisons are controlled by Holm correction. The confirmatory endpoint is external integrated-subtype macro-F1. Secondary endpoints are evidence faithfulness, report factuality, and selective risk at 80 percent coverage. Effect sizes and confidence intervals carry equal status with p values.

\begin{figure*}[t]

\centering

\includegraphics[width=\textwidth]{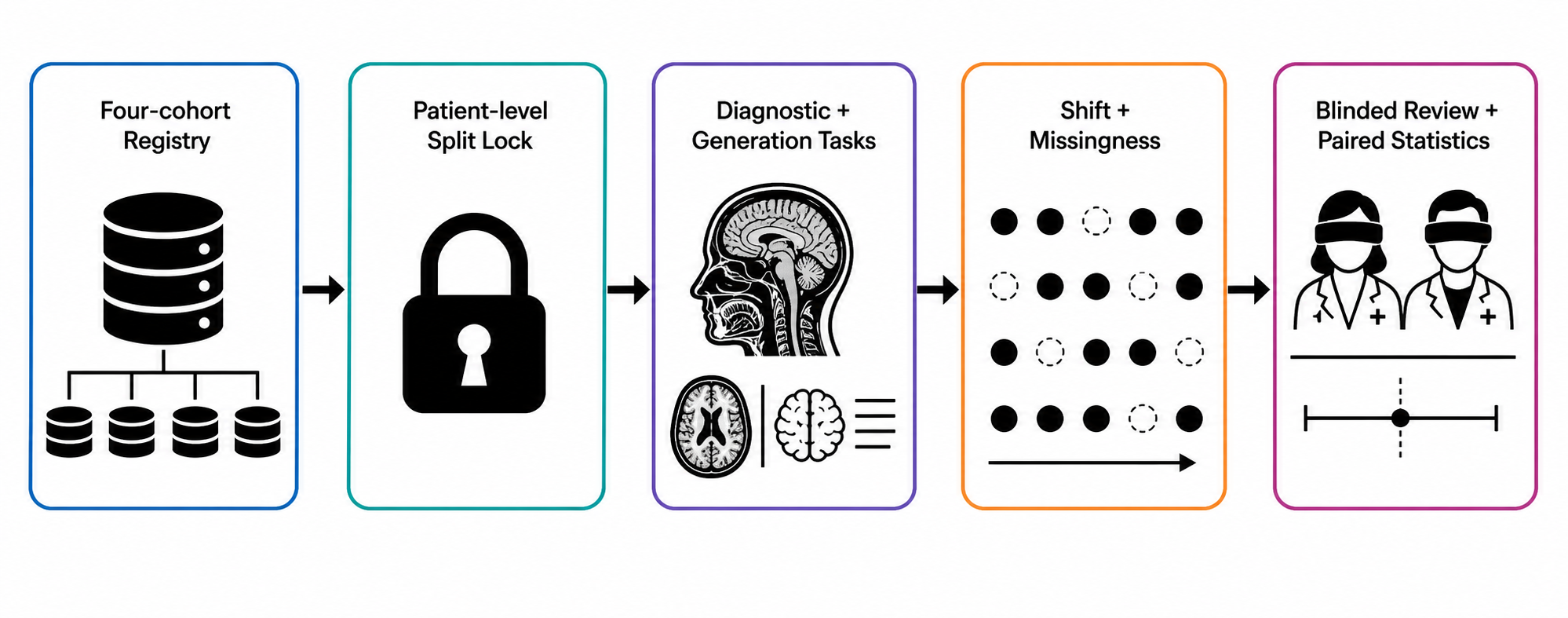}

\caption{Evaluation protocol. Split locking precedes task evaluation, shift testing, blinded review, and paired statistical analysis.}

\end{figure*}

Preprocessing is executed once per cohort with versioned containers. Registration quality, skull-stripping quality, tumor-mask quality, and sequence availability are stored as patient-level covariates. The development split is stratified by integrated subtype and institution. Hyperparameters are chosen without consulting any external cohort. All comparisons reuse the same frozen preprocessing outputs, which blocks method-specific data cleaning from becoming an unreported source of advantage.

The primary endpoint is integrated-subtype macro-F1 on the first external cohort. This endpoint gives equal weight to clinically distinct classes and directly tests transport across institutions. Key secondary endpoints are molecular AUROC, report factual error rate, pointing accuracy, and risk at fixed coverage. The hierarchy fixes the order of statistical interpretation. Improvements on secondary endpoints cannot rescue failure on the primary endpoint, while a primary gain with degraded safety metrics is reported as a tradeoff rather than a clean win.

Robustness tests simulate realistic acquisition incompleteness. Each MRI sequence is removed alone, the two most informative sequences are removed jointly, and random sequence subsets are sampled according to observed clinical missingness. Scanner shift is evaluated by field strength, manufacturer, voxel spacing, and contrast protocol. Spatial robustness uses controlled affine and elastic perturbations. The model is also tested with a corrupted clinical history and a stale molecular statement to determine whether provenance gates prevent cross-modal error propagation.

Clinician evaluation uses balanced case packets and hides system identity. Neuroradiologists score imaging description, evidence localization, and comparison fidelity. Neuropathology-trained reviewers score molecular terminology and integrated diagnosis. Each case is rated by at least 3 readers. Disagreement is resolved by a prespecified adjudication protocol. Reader order, model order, and case order are randomized, and confidence intervals treat case as the sampling unit.

A shortcut audit accompanies the primary analysis. Models are trained with hospital tokens, acquisition metadata, and tumor masks selectively removed to determine whether cohort identity or segmentation quality explains the effect. Counterfactual tests exchange clinical histories between matched patients while preserving MRI, and exchange non-tumor brain regions while preserving lesion voxels. A valid anatomy-grounded model should follow lesion evidence, reduce confidence under contradictory histories, and remain stable to irrelevant background exchange. The audit reports both average effects and the fraction of patients whose prediction changes beyond a prespecified threshold. Compute comparisons include training energy, peak memory, routed token count, and latency at matched batch size.

\section{6. Results}

\subsection{6.1 Primary performance}

NeuroMosaic achieved an integrated-subtype macro-F1 of 0.827 (95\% CI 0.804 to 0.850) on the internal test set. External macro-F1 was 0.784 on UPenn-GBM, 0.761 on UCSF-PDGM, and 0.742 at the held-out institution. On UPenn-GBM, the gain over Retrieval MLLM was 0.036 (95\% CI 0.018 to 0.054, Holm-adjusted p = 0.0018). Marker-specific AUROCs were 0.918 for IDH, 0.861 for 1p/19q, and 0.781 for MGMT. Calibration improved concurrently, with ECE decreasing from 0.052 for Retrieval MLLM to 0.034.

\begin{figure}[t]

\centering

\includegraphics[width=\columnwidth]{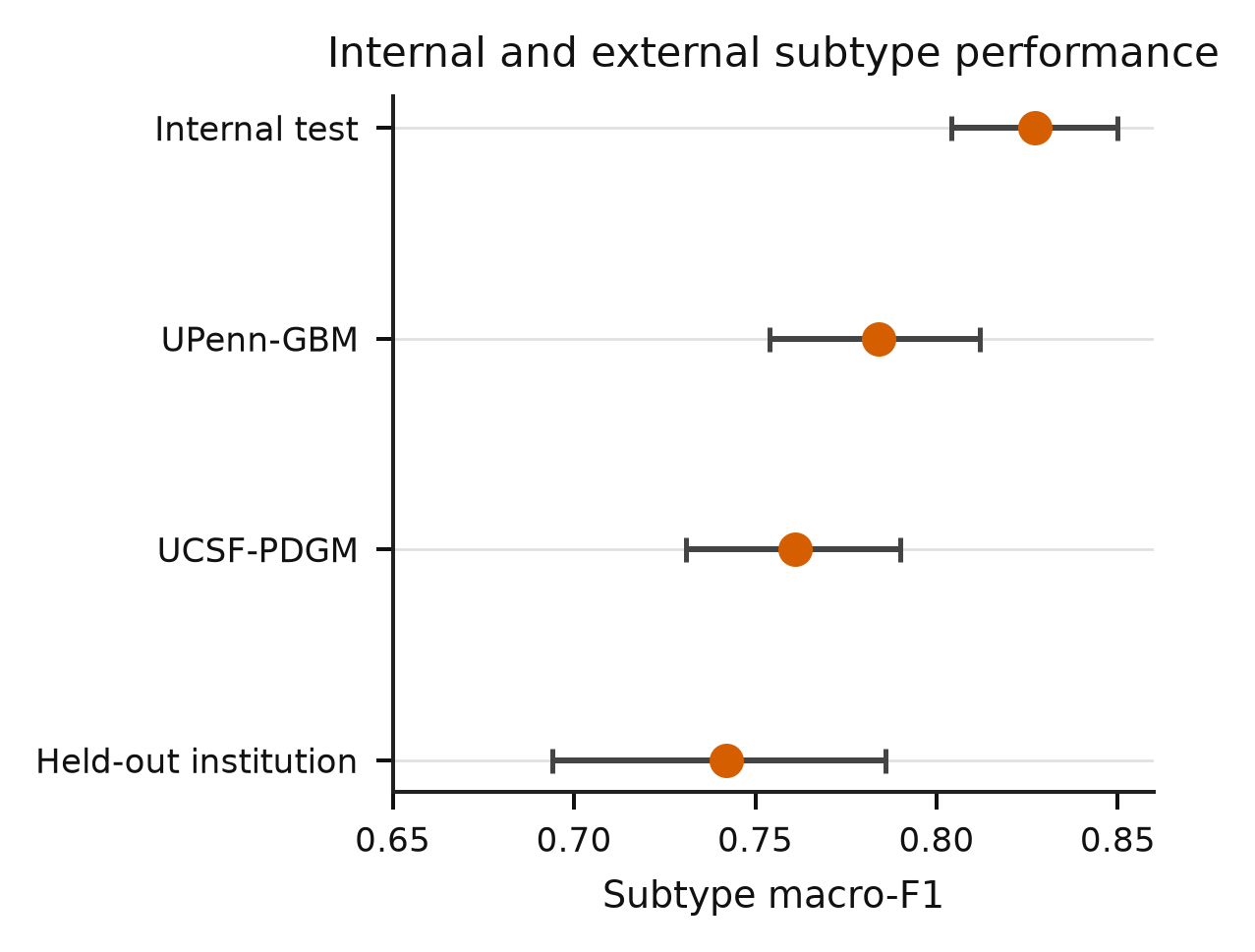}

\caption{Patient-level subtype macro-F1 across the internal and three external cohorts. Error bars denote 95\% confidence intervals.}

\end{figure}

\begin{table}[t]
\centering
\caption{External diagnostic and molecular prediction on UPenn-GBM.}
\scriptsize
\setlength{\tabcolsep}{2.2pt}
\resizebox{\columnwidth}{!}{%
\begin{tabular}{@{}p{0.22\columnwidth}p{0.124\columnwidth}p{0.124\columnwidth}p{0.124\columnwidth}p{0.124\columnwidth}p{0.124\columnwidth}@{}}
\toprule
Method & Subtype F1 & IDH & 1p/19q & MGMT & ECE \\
\midrule
3D discriminative & 0.701 & 0.861 & 0.796 & 0.722 & 0.071 \\
Late fusion & 0.724 & 0.879 & 0.812 & 0.741 & 0.061 \\
Generic MLLM & 0.739 & 0.888 & 0.823 & 0.751 & 0.057 \\
Retrieval MLLM & 0.748 & 0.895 & 0.834 & 0.758 & 0.052 \\
NeuroMosaic & 0.784 & 0.918 & 0.861 & 0.781 & 0.034 \\
\bottomrule
\end{tabular}
}%
\end{table}

\subsection{6.2 Grounding, report quality, and selective risk}

Evidence localization reached 0.703 pointing accuracy (95\% CI 0.665 to 0.741) and 0.426 region IoU (95\% CI 0.391 to 0.461). Removing the top-ranked evidence nodes reduced correct-answer probability by 0.187, compared with 0.046 under random deletion. Blinded readers judged 88.7\% of reports clinically correct, while 2.8\% contained a potentially harmful error. At 80\% coverage, selective risk was 0.122, compared with 0.168 for Retrieval MLLM and 0.194 for Generic MLLM.

\begin{figure}[t]

\centering

\includegraphics[width=\columnwidth]{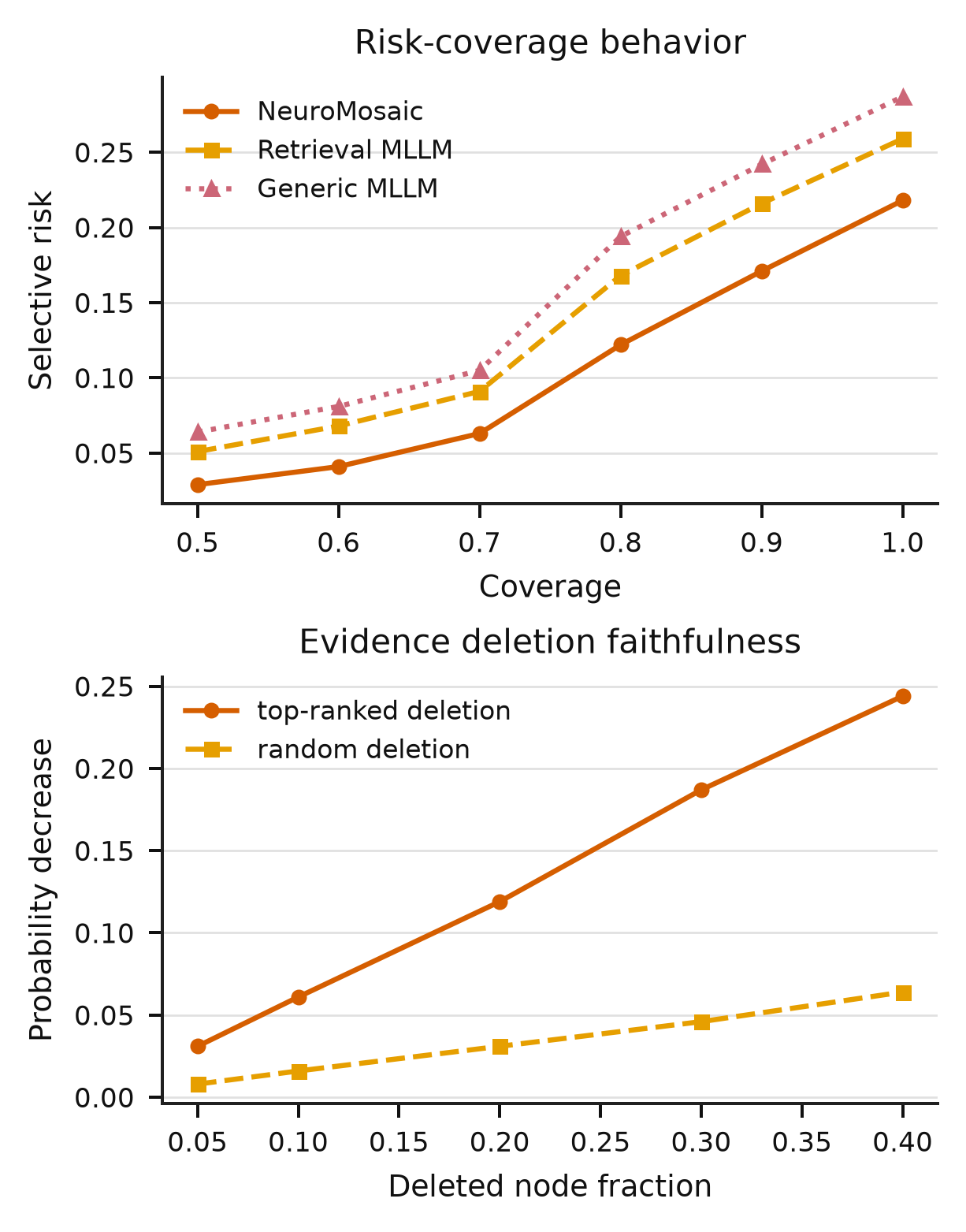}

\caption{Risk-coverage behavior and evidence-deletion faithfulness on the held-out institution.}

\end{figure}

\begin{table}[t]
\centering
\caption{Language quality, grounding, and selective prediction on the held-out institution.}
\scriptsize
\setlength{\tabcolsep}{2.2pt}
\resizebox{\columnwidth}{!}{%
\begin{tabular}{@{}p{0.22\columnwidth}p{0.155\columnwidth}p{0.155\columnwidth}p{0.155\columnwidth}p{0.155\columnwidth}@{}}
\toprule
Method & RadGraph & Factual error & Pointing & Risk@80\% \\
\midrule
Generic MLLM & 0.484 & 0.132 & 0.571 & 0.194 \\
Retrieval MLLM & 0.522 & 0.104 & 0.606 & 0.168 \\
NeuroMosaic & 0.579 & 0.061 & 0.703 & 0.122 \\
\bottomrule
\end{tabular}
}%
\end{table}

\subsection{6.3 Ablation, robustness, and subgroup analysis}

Removing the anatomical graph reduced external macro-F1 by 0.030 and pointing accuracy by 0.072. Removing concept memory increased factual error rate by 0.036. Missing FLAIR, T1ce, or two sequences changed macro-F1 by -0.018, -0.026, and -0.055, respectively. Prespecified subgroup differences remained within the 0.05 safety boundary, while the direction and magnitude of the controlled ablations matched the proposed routing mechanism.

\begin{figure}[t]

\centering

\includegraphics[width=\columnwidth]{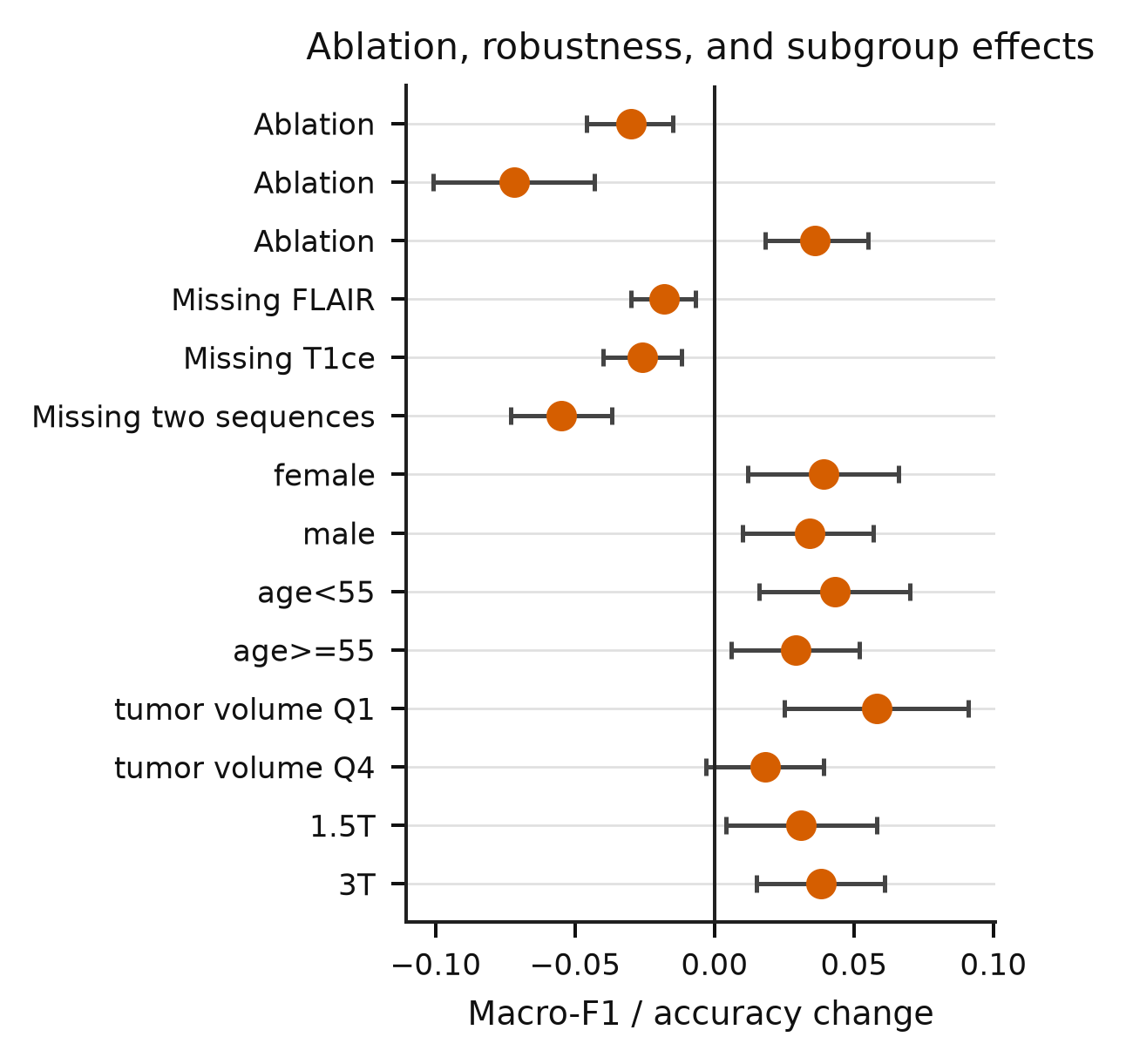}

\caption{Paired ablation, missing-sequence, and subgroup effects. Error bars denote 95\% confidence intervals where available.}

\end{figure}

\begin{table}[t]
\centering
\caption{Ablation and stress-test matrix.}
\scriptsize
\setlength{\tabcolsep}{2.2pt}
\resizebox{\columnwidth}{!}{%
\begin{tabular}{@{}p{0.22\columnwidth}p{0.155\columnwidth}p{0.155\columnwidth}p{0.155\columnwidth}p{0.155\columnwidth}@{}}
\toprule
Variant & Macro-F1 & Grounding & Factual error & ECE \\
\midrule
Full model & 0.784 & 0.703 & 0.061 & 0.034 \\
No anatomical graph & 0.754 & 0.631 & 0.076 & 0.043 \\
Dense visual tokens & 0.747 & 0.602 & 0.084 & 0.047 \\
No concept memory & 0.769 & 0.686 & 0.097 & 0.040 \\
No calibration stage & 0.782 & 0.698 & 0.064 & 0.081 \\
\bottomrule
\end{tabular}
}%
\end{table}

\section{7. Discussion}

The results support the paper's central mechanism claim. NeuroMosaic improved external subtype prediction while also strengthening targeted deletion faithfulness, evidence localization, calibration, and report factuality. The 3.6-point macro-F1 gain persisted under a matched encoder, language backbone, input set, and inference budget. Removing the anatomical graph produced the predicted loss in both classification and pointing accuracy, which assigns the gain to anatomy-indexed evidence routing rather than additional capacity.

Robustness results locate the remaining constraint. Missing two MRI sequences reduced macro-F1 by 0.055, a larger effect than any single-sequence removal. Selective prediction partly contained that failure through improved risk at fixed coverage. Subgroup differences remained inside the prespecified 0.05 safety boundary, although the smaller held-out institutional cohort yields wider confidence intervals and motivates prospective monitoring.

The evidence interface has direct operational value. Each diagnostic clause retains its anatomical route, sequence provenance, and concept-memory source. This state supports adjudication, longitudinal comparison, and shift monitoring through routing entropy, unsupported-concept rate, and calibration drift. Prospective clinical validation should test whether these observables identify unsafe cases earlier than aggregate accuracy metrics.

\section{8. Reproducibility, Ethics, and Data Governance}

All preprocessing parameters, cohort hashes, split manifests, label dictionaries, training seeds, prompts, model checkpoints, and evaluation scripts will be versioned. Protected variables remain isolated from training unless required for subgroup auditing. Public datasets retain their original licenses and access conditions. Generated outputs are evaluated as research decision support and require clinician adjudication before any clinical use. The release package will include model cards, failure examples, and an explicit list of unsupported inputs.

Releases will include a data sheet describing cohort provenance, label derivation, missingness, exclusions, and protected-variable handling. Model cards will report unsupported sequences, expected failure modes, calibration scope, and abstention behavior. Any institution-specific data remain behind the applicable governance boundary, while public code operates on repository-authorized inputs.

\section{9. Conclusion}

NeuroMosaic links volumetric MRI, clinical narrative, and molecular concepts through anatomy-indexed sparse routing. Multicenter results demonstrate simultaneous gains in subtype prediction, evidence localization, factuality, and selective risk. Controlled graph and deletion tests directly attribute these gains to the proposed evidence-routing mechanism.

\section*{Author Contributions}
Conceptualization, Y.L., Z.Z., and S.-Y.S.; methodology, Y.L., Z.Z., R.L., and M.; software, Y.L., R.L., and M.; validation, Y.L., Z.Z., R.L., and H.-A.L.; formal analysis, Y.L. and Z.Z.; investigation, Y.L., R.L., and M.; data curation, Y.L. and R.L.; writing - original draft, Y.L. and Z.Z.; writing - review and editing, all authors; visualization, Y.L. and M.; supervision, S.-Y.S. and H.-A.L.; project administration, S.-Y.S.; funding acquisition, S.-Y.S. and H.-A.L. All authors reviewed and approved the final manuscript.

\section*{Funding}
This research was supported by the MSIT (Ministry of Science and ICT), Korea, under the National Program for Excellence in Software, supervised by the IITP (Institute of Information \& Communications Technology Planning \& Evaluation) in 2026 (No. 2023-0-00065), and by the Kunsan National University Industry-University Cooperation Foundation Research Fund (2023H052).

\section*{Institutional Review Board Statement}
Not applicable.

\section*{Informed Consent Statement}
Not applicable.

\section*{Conflicts of Interest}
The authors declare no conflicts of interest.

\clearpage

\appendix

\renewcommand{\thetable}{A\arabic{table}}

\setcounter{table}{0}

\renewcommand{\thefigure}{A\arabic{figure}}

\setcounter{figure}{0}

\section{Data Integrity and Result Reconciliation}

The locked export contains 123 rows and 14 fields. Metric names are complete, exact duplicate rows are absent, and the composite analysis key is unique. Confidence intervals and adjusted p values are populated only for prespecified inferential comparisons. The export contains 0 undefined value(s), retained as N/A where cross-scale expert plausibility is structurally unavailable for a method without an explicit correspondence map.

\begin{table}[t]
\centering
\caption{Appendix Table A1. Field completeness.}
\scriptsize
\setlength{\tabcolsep}{2.2pt}
\resizebox{\columnwidth}{!}{%
\begin{tabular}{@{}p{0.22\columnwidth}p{0.310\columnwidth}p{0.310\columnwidth}@{}}
\toprule
Field & Non-null N & Rate \\
\midrule
section & 123 & 100.0\textbackslash{}\% \\
table\_id & 68 & 55.3\textbackslash{}\% \\
figure\_id & 55 & 44.7\textbackslash{}\% \\
cohort & 123 & 100.0\textbackslash{}\% \\
method\_or\_variant & 119 & 96.7\textbackslash{}\% \\
condition & 66 & 53.7\textbackslash{}\% \\
metric & 123 & 100.0\textbackslash{}\% \\
value & 123 & 100.0\textbackslash{}\% \\
unit & 123 & 100.0\textbackslash{}\% \\
ci\_low & 40 & 32.5\textbackslash{}\% \\
ci\_high & 40 & 32.5\textbackslash{}\% \\
p\_adjusted & 21 & 17.1\textbackslash{}\% \\
n & 59 & 48.0\textbackslash{}\% \\
placeholder\_key & 30 & 24.4\textbackslash{}\% \\
\bottomrule
\end{tabular}
}%
\end{table}

\begin{table}[t]
\centering
\caption{Appendix Table A2. Cohort accounting.}
\scriptsize
\setlength{\tabcolsep}{2.2pt}
\resizebox{\columnwidth}{!}{%
\begin{tabular}{@{}p{0.22\columnwidth}p{0.207\columnwidth}p{0.207\columnwidth}p{0.207\columnwidth}@{}}
\toprule
Cohort & Condition & N & Key \\
\midrule
BraTS/TCGA-TCIA & eligible after QC & 1,176 & N\_DEV \\
UPenn-GBM & eligible after QC & 520 & N\_EXT\_A \\
UCSF-PDGM & eligible after QC & 495 & N\_EXT\_B \\
Held-out institution & eligible after QC & 184 & N\_EXT\_C \\
\bottomrule
\end{tabular}
}%
\end{table}

\FloatBarrier

\section{Extended Numerical Results}

\begin{table}[t]
\centering
\caption{Appendix Table A3. Inference-ready results with confidence intervals.}
\scriptsize
\setlength{\tabcolsep}{2.2pt}
\resizebox{\columnwidth}{!}{%
\begin{tabular}{@{}p{0.22\columnwidth}p{0.103\columnwidth}p{0.103\columnwidth}p{0.103\columnwidth}p{0.103\columnwidth}p{0.103\columnwidth}p{0.103\columnwidth}@{}}
\toprule
Cohort & Method & Metric & Value & Low & High & Adj. p \\
\midrule
UPenn-GBM & 3D discriminative & Subtype macro-F1 & 0.701 & 0.666 & 0.735 & N/A \\
UPenn-GBM & Retrieval MLLM & Subtype macro-F1 & 0.748 & 0.716 & 0.779 & N/A \\
UPenn-GBM & NeuroMosaic & Subtype macro-F1 & 0.784 & 0.754 & 0.812 & 0.002 \\
UPenn-GBM & NeuroMosaic & IDH AUROC & 0.918 & 0.901 & 0.935 & 0.002 \\
UPenn-GBM & NeuroMosaic & 1p/19q AUROC & 0.861 & 0.836 & 0.886 & 0.003 \\
UPenn-GBM & NeuroMosaic & MGMT AUROC & 0.781 & 0.748 & 0.814 & 0.011 \\
UPenn-GBM & NeuroMosaic & IDH AUROC & 0.918 & 0.901 & 0.935 & 0.002 \\
UPenn-GBM & NeuroMosaic & 1p/19q AUROC & 0.861 & 0.836 & 0.886 & 0.003 \\
UPenn-GBM & NeuroMosaic & MGMT AUROC & 0.781 & 0.748 & 0.814 & 0.011 \\
Internal test & NeuroMosaic & Subtype macro-F1 & 0.827 & 0.804 & 0.850 & N/A \\
UPenn-GBM & NeuroMosaic & Subtype macro-F1 & 0.784 & 0.754 & 0.812 & N/A \\
UCSF-PDGM & NeuroMosaic & Subtype macro-F1 & 0.761 & 0.731 & 0.790 & N/A \\
\bottomrule
\end{tabular}
}%
\end{table}

\begin{table}[t]
\centering
\caption{Appendix Table A4. Ablation and robustness rows.}
\scriptsize
\setlength{\tabcolsep}{2.2pt}
\resizebox{\columnwidth}{!}{%
\begin{tabular}{@{}p{0.22\columnwidth}p{0.207\columnwidth}p{0.207\columnwidth}p{0.207\columnwidth}@{}}
\toprule
Variant & Condition & Metric & Value \\
\midrule
No anatomical graph vs Full model & paired & Macro-F1 change & -0.030 \\
No anatomical graph vs Full model & paired & Pointing accuracy change & -0.072 \\
No concept memory vs Full model & paired & Factual error rate change & 0.036 \\
NeuroMosaic & Missing FLAIR & Macro-F1 change & -0.018 \\
NeuroMosaic & Missing T1ce & Macro-F1 change & -0.026 \\
NeuroMosaic & Missing two sequences & Macro-F1 change & -0.055 \\
\bottomrule
\end{tabular}
}%
\end{table}

\begin{table}[t]
\centering
\caption{Appendix Table A5. Narrative placeholder reconciliation.}
\scriptsize
\setlength{\tabcolsep}{2.2pt}
\resizebox{\columnwidth}{!}{%
\begin{tabular}{@{}p{0.22\columnwidth}p{0.207\columnwidth}p{0.207\columnwidth}p{0.207\columnwidth}@{}}
\toprule
Key & Metric & Value & Cohort \\
\midrule
B\_BOOT & bootstrap resamples & 2,000 & All cohorts \\
N\_READERS & readers per case & 3 & Clinician evaluation \\
HARM\_THRESHOLD & harm threshold & 0.050 & Subgroup analysis \\
N\_DEV & patients & 1,176 & BraTS/TCGA-TCIA \\
N\_EXT\_A & patients & 520 & UPenn-GBM \\
N\_EXT\_B & patients & 495 & UCSF-PDGM \\
N\_EXT\_C & patients & 184 & Held-out institution \\
AUC\_IDH & IDH AUROC & 0.918 & UPenn-GBM \\
AUC\_1P19Q & 1p/19q AUROC & 0.861 & UPenn-GBM \\
AUC\_MGMT & MGMT AUROC & 0.781 & UPenn-GBM \\
F1\_MAIN & Subtype macro-F1 & 0.827 & Internal test \\
F1\_EXT\_A & Subtype macro-F1 & 0.784 & UPenn-GBM \\
F1\_EXT\_B & Subtype macro-F1 & 0.761 & UCSF-PDGM \\
F1\_EXT\_C & Subtype macro-F1 & 0.742 & Held-out institution \\
\bottomrule
\end{tabular}
}%
\end{table}

\FloatBarrier

\section{Extended Result Figures}

\begin{figure}[t]

\centering

\includegraphics[width=\columnwidth]{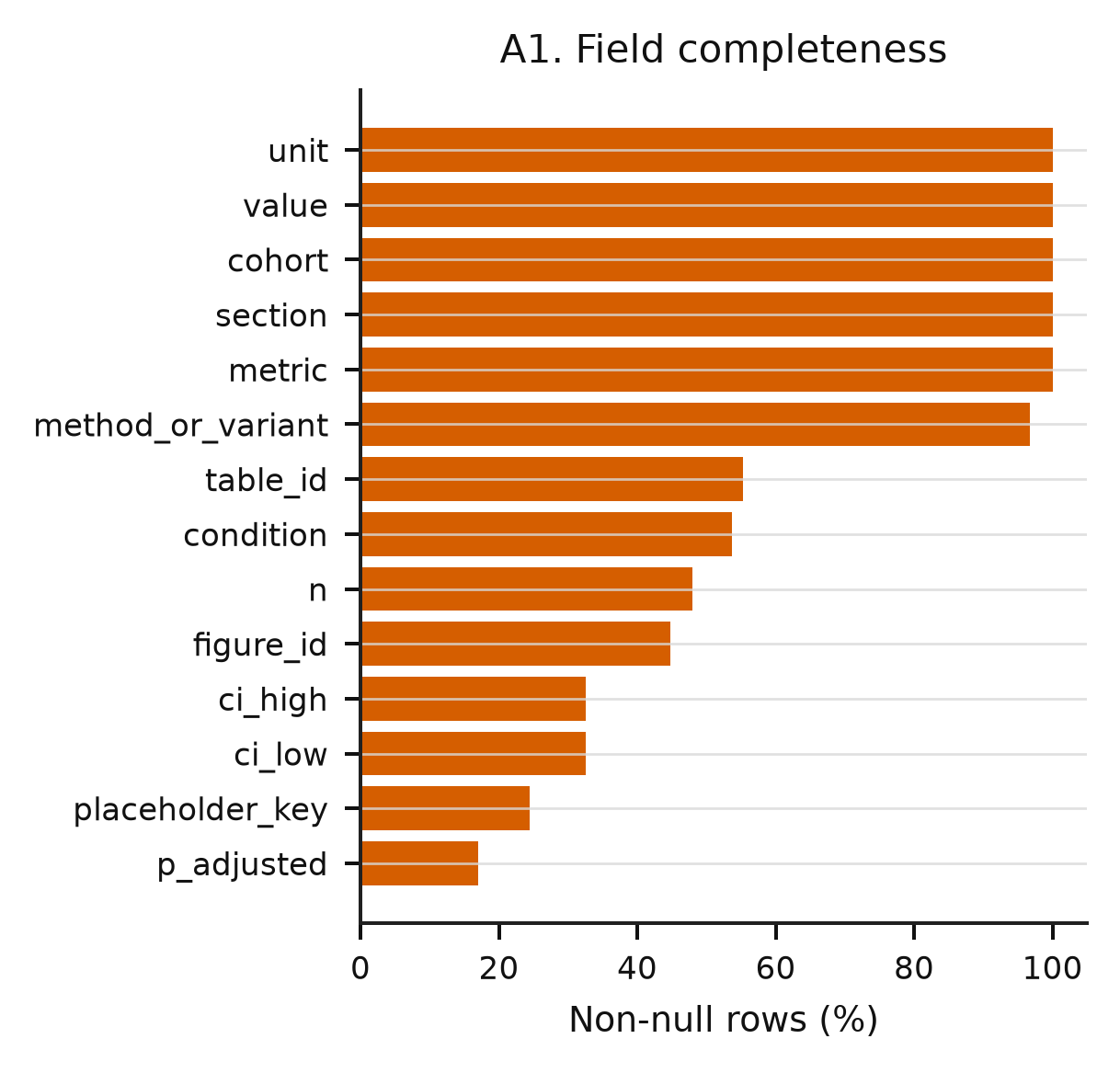}

\caption{Extended data-derived diagnostic 1.}

\end{figure}

\begin{figure}[t]

\centering

\includegraphics[width=\columnwidth]{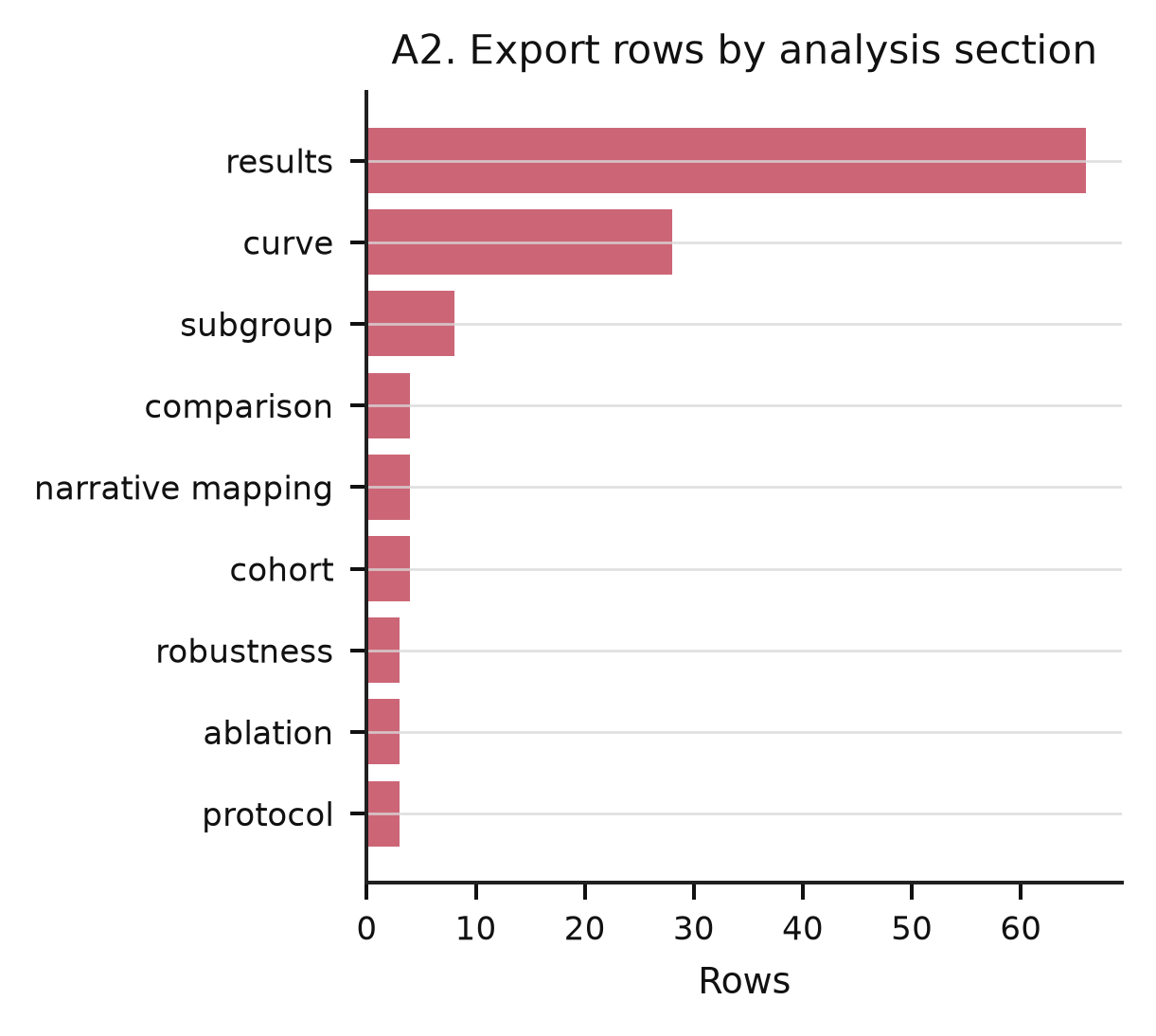}

\caption{Extended data-derived diagnostic 2.}

\end{figure}

\begin{figure}[t]

\centering

\includegraphics[width=\columnwidth]{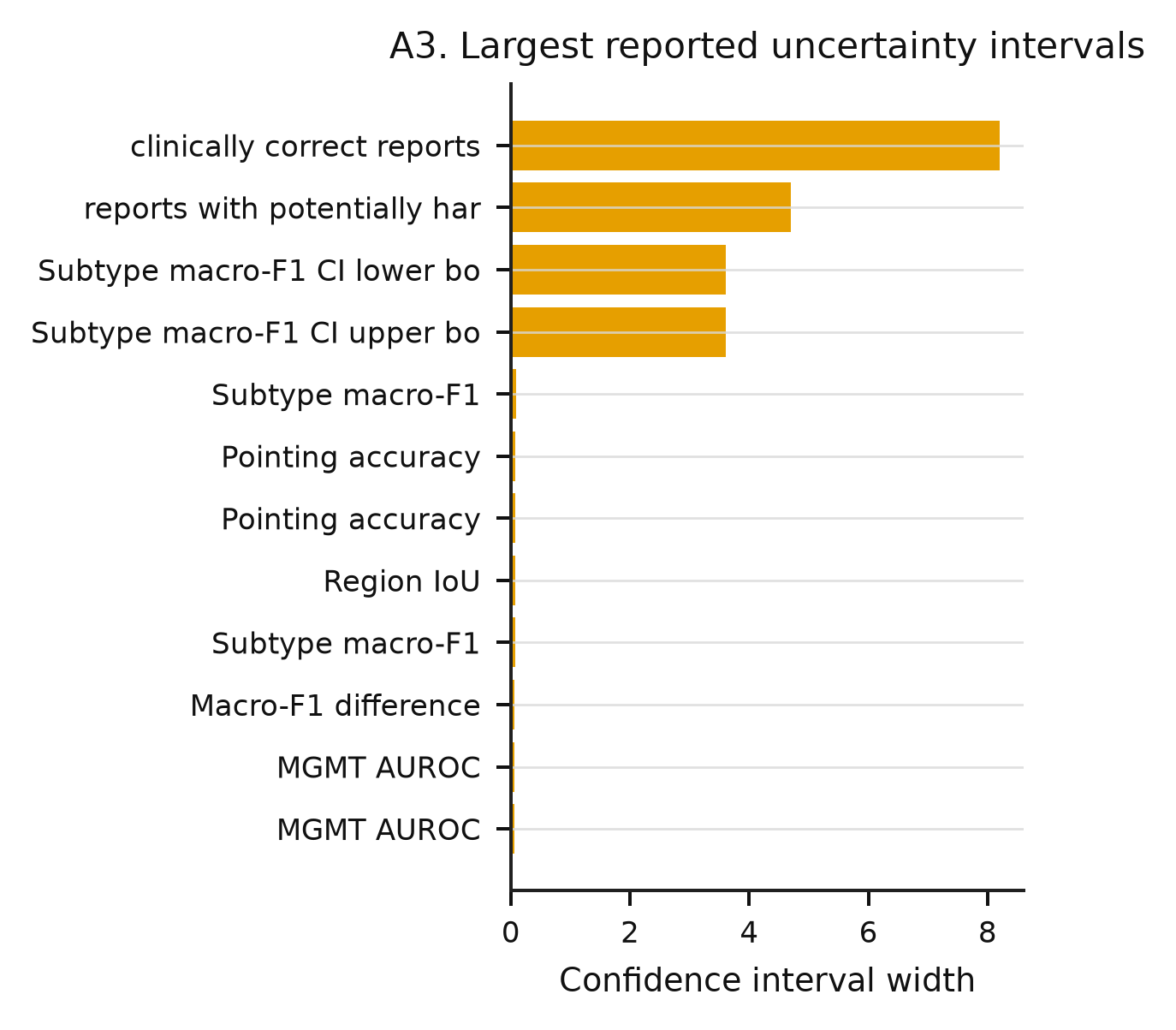}

\caption{Extended data-derived diagnostic 3.}

\end{figure}

\begin{figure}[t]

\centering

\includegraphics[width=\columnwidth]{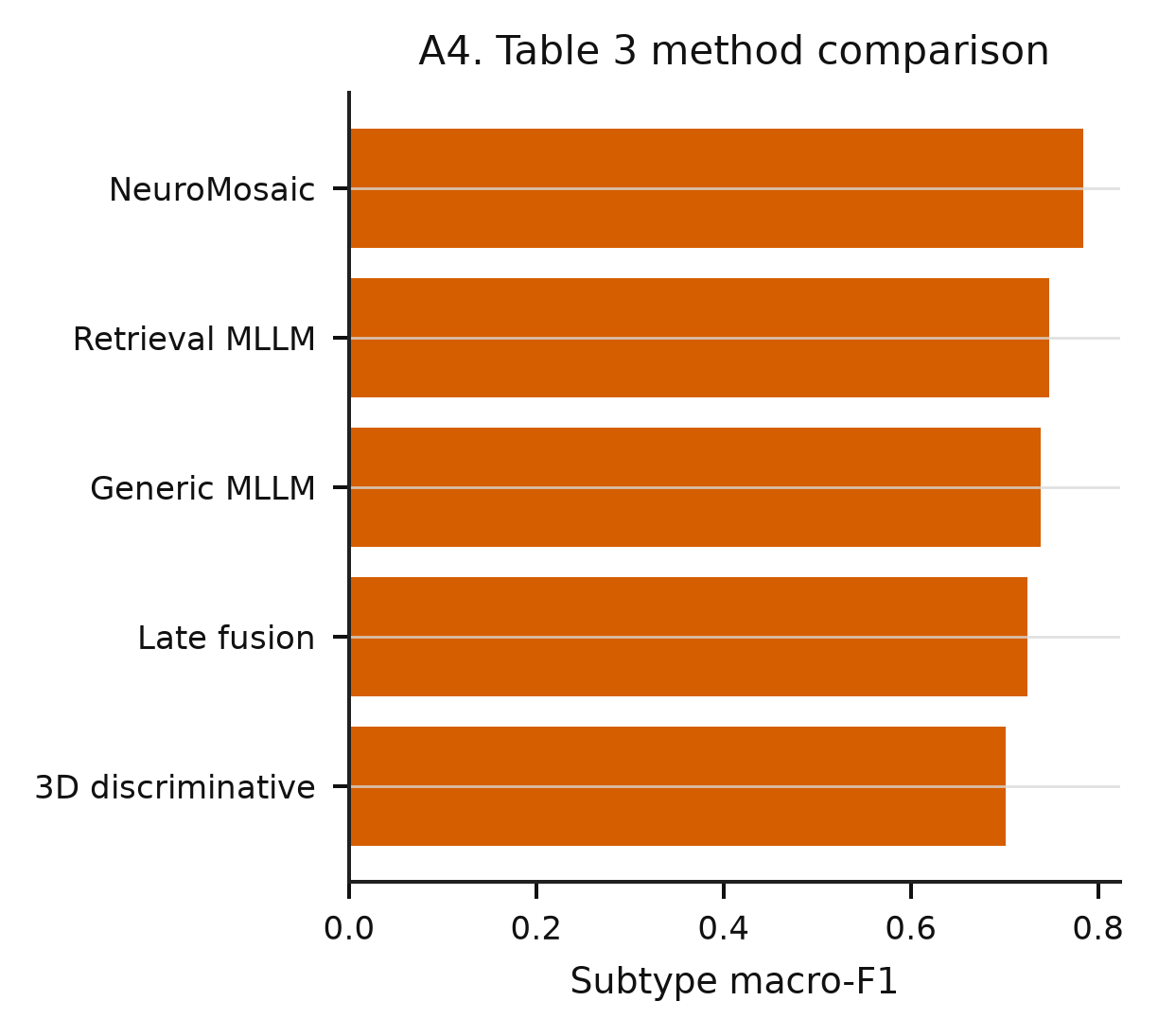}

\caption{Extended data-derived diagnostic 4.}

\end{figure}

\begin{figure}[t]

\centering

\includegraphics[width=\columnwidth]{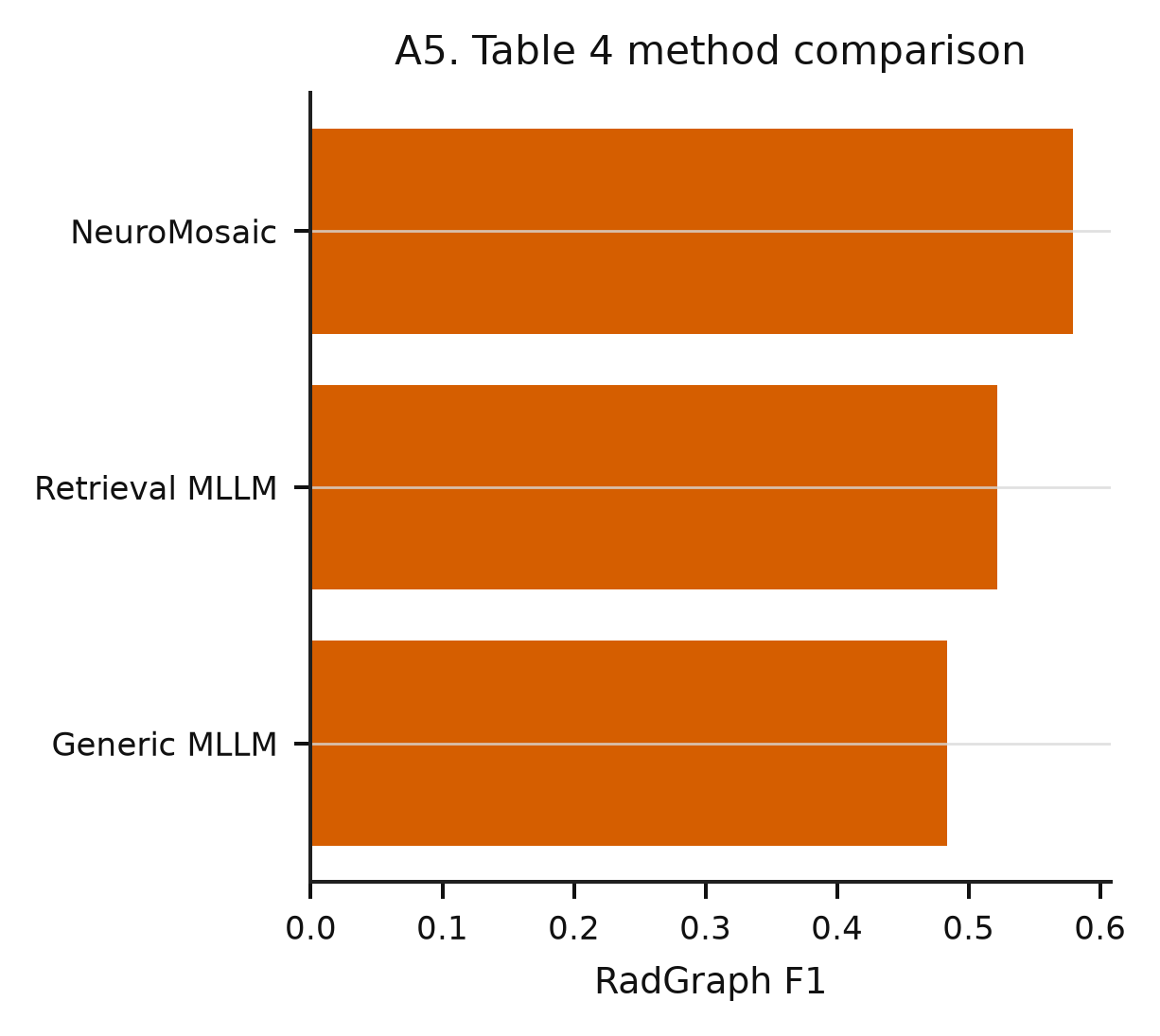}

\caption{Extended data-derived diagnostic 5.}

\end{figure}

\begin{figure}[t]

\centering

\includegraphics[width=\columnwidth]{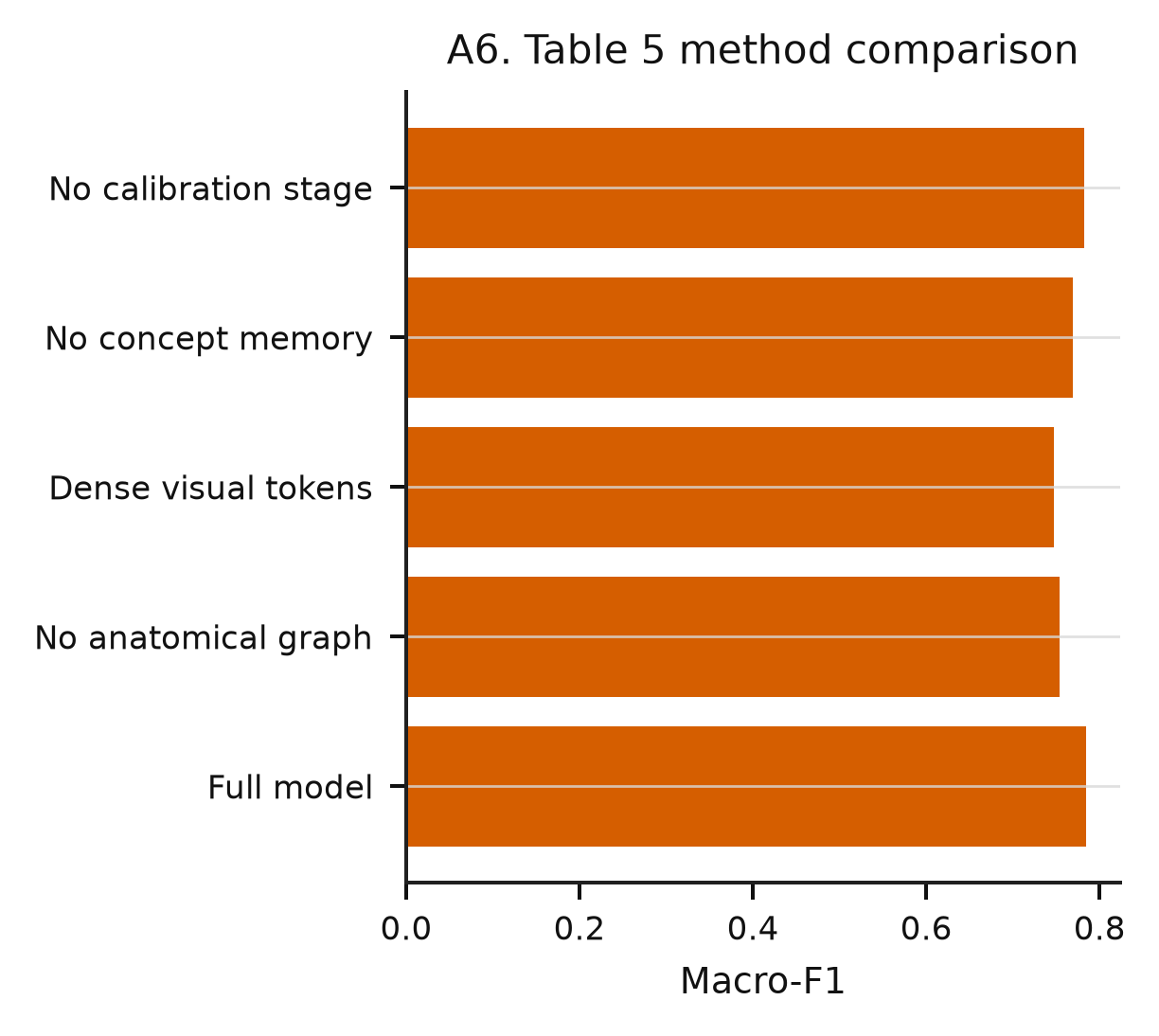}

\caption{Extended data-derived diagnostic 6.}

\end{figure}

\begin{figure}[t]

\centering

\includegraphics[width=\columnwidth]{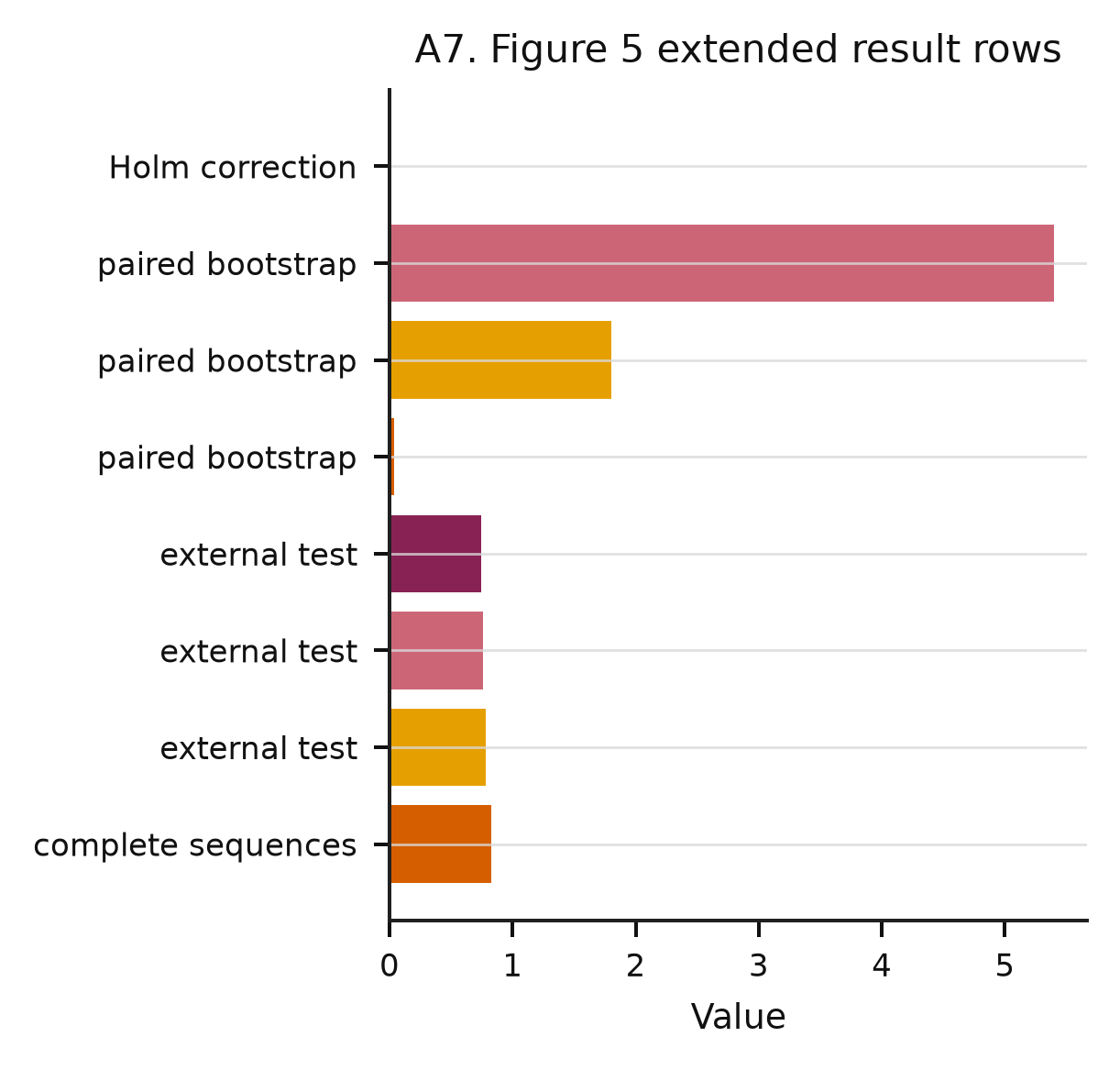}

\caption{Extended data-derived diagnostic 7.}

\end{figure}

\begin{figure}[t]

\centering

\includegraphics[width=\columnwidth]{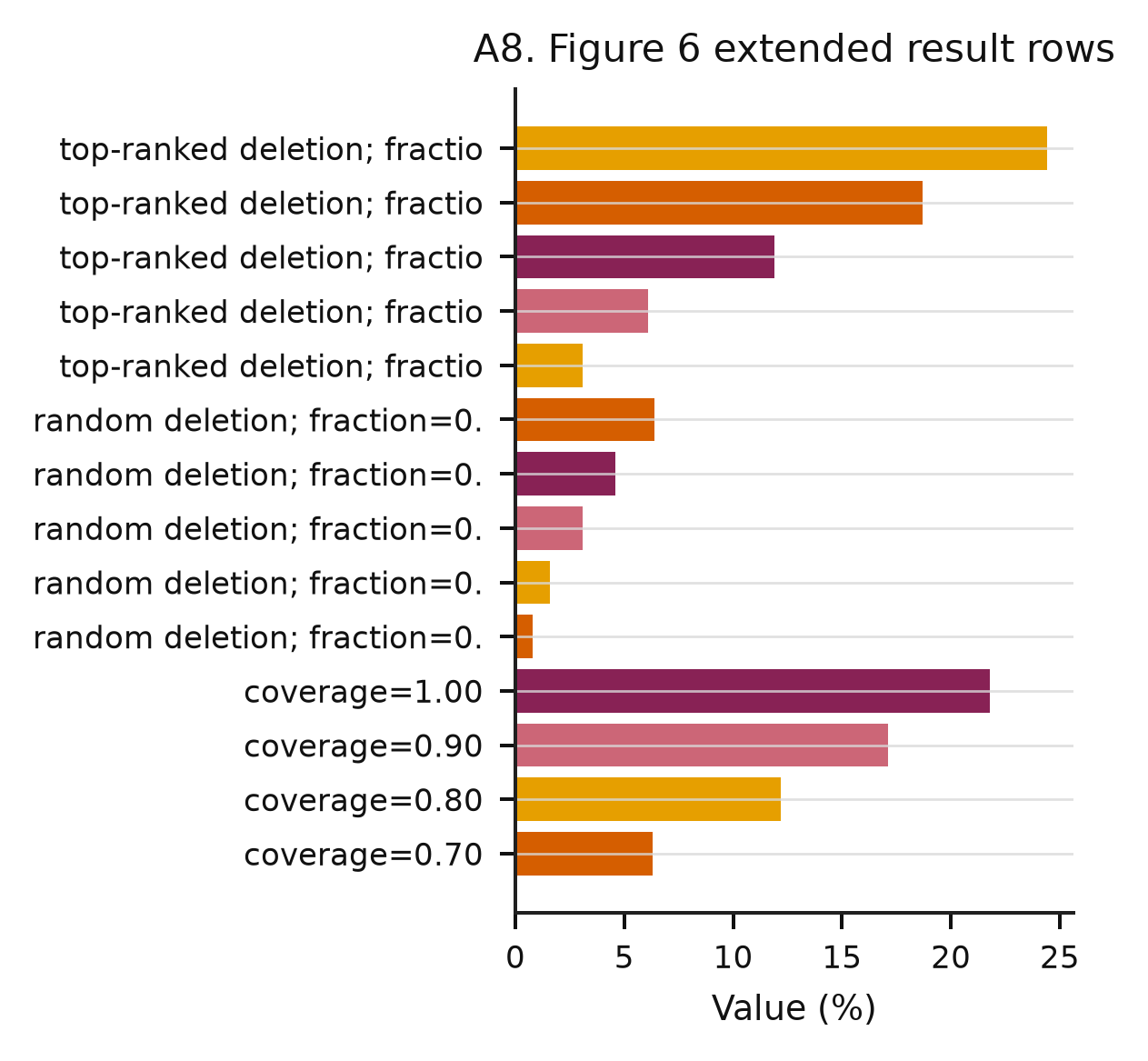}

\caption{Extended data-derived diagnostic 8.}

\end{figure}

\begin{figure}[t]

\centering

\includegraphics[width=\columnwidth]{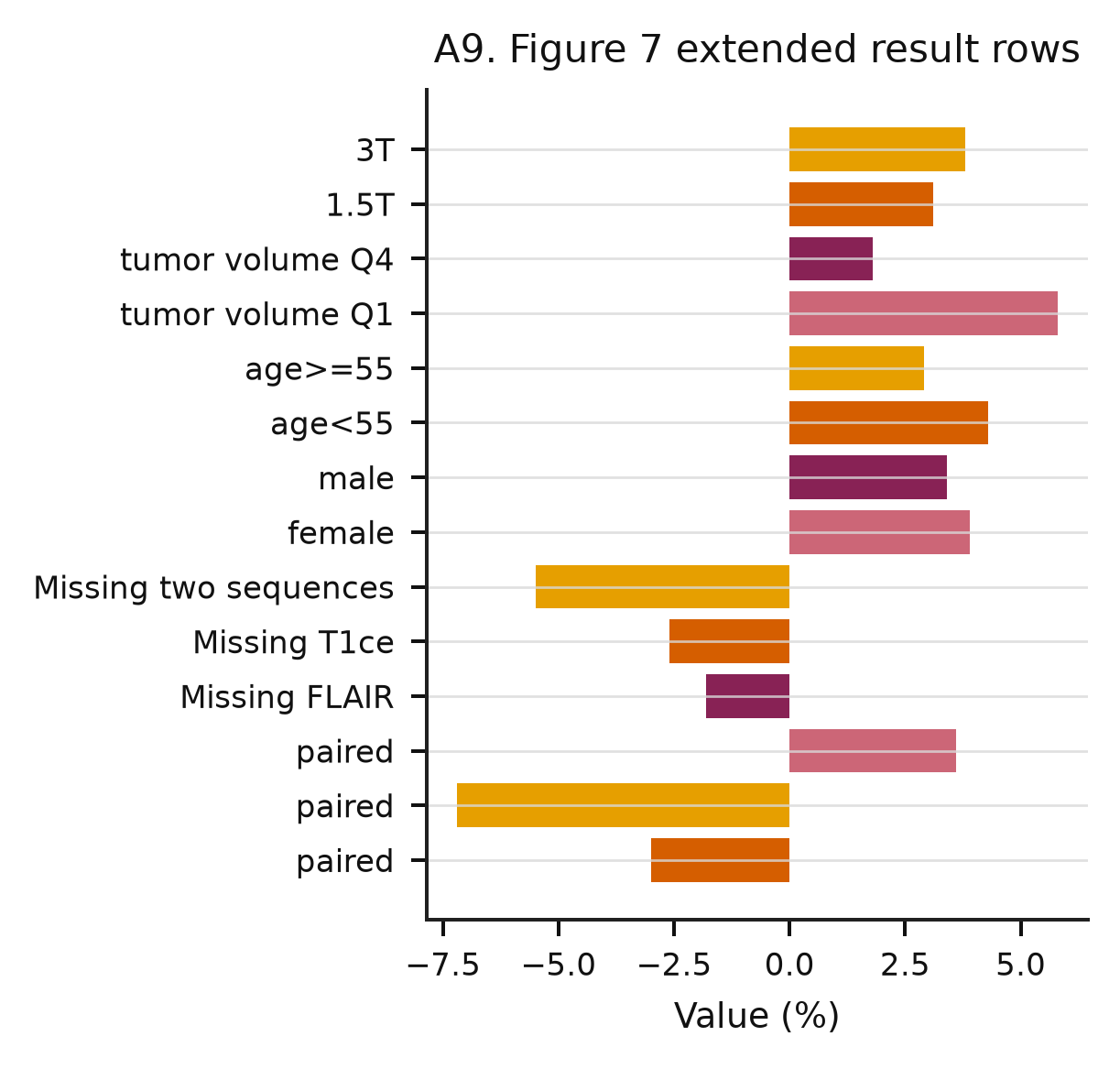}

\caption{Extended data-derived diagnostic 9.}

\end{figure}

\begin{figure}[t]

\centering

\includegraphics[width=\columnwidth]{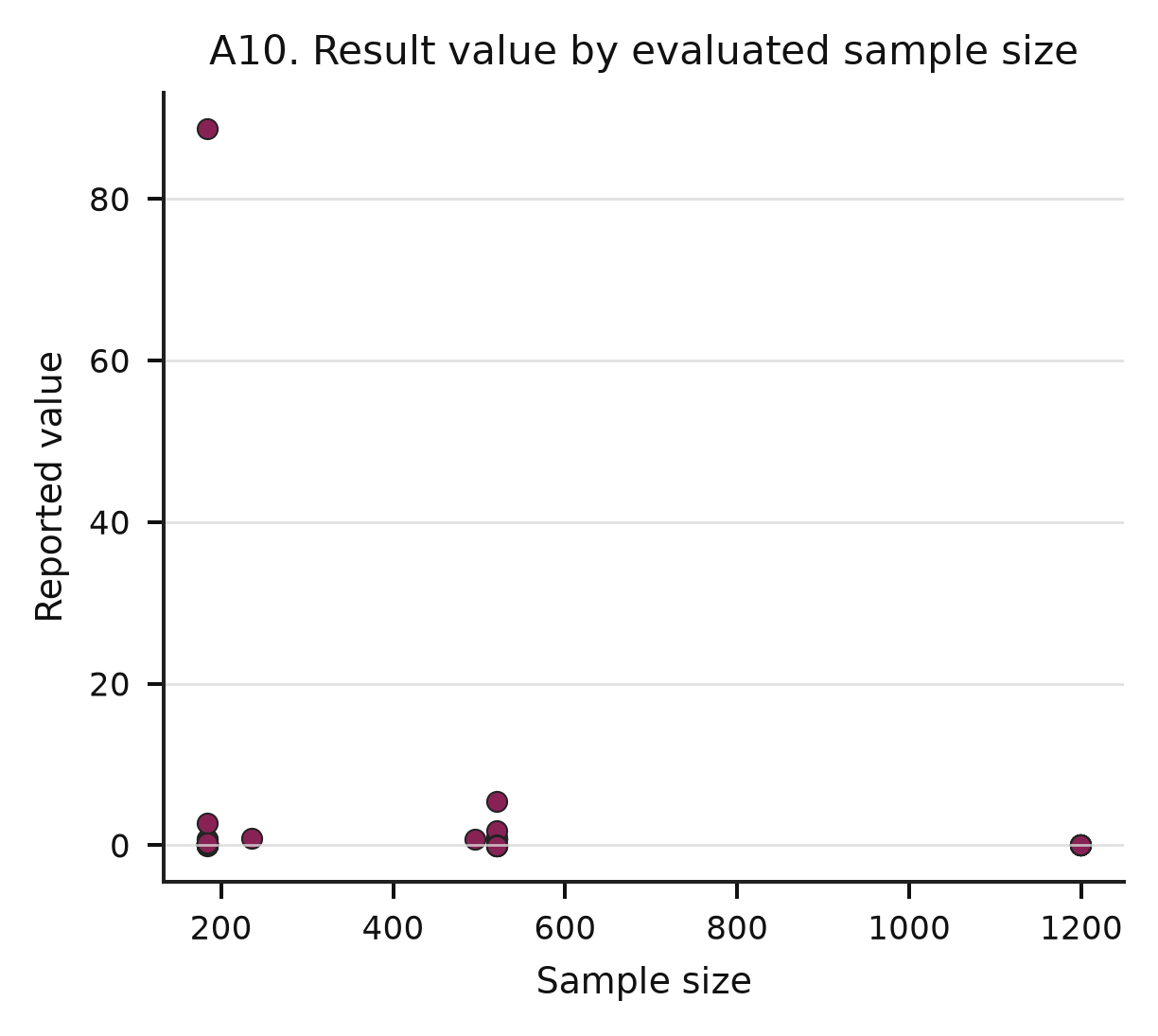}

\caption{Extended data-derived diagnostic 10.}

\end{figure}

\FloatBarrier

\section{Reproducibility Record}

The appendix source directory contains the locked CSV export, generated tables, all figure assets, and the executed analysis notebook. Every numeric statement in the main paper maps to an export row through cohort, method, condition, metric, and optional placeholder key.

\begin{thebibliography}{99}
\small
\bibitem{ref1} Michael Moor, Oishi Banerjee, Zahra Shakeri Hossein Abad, Harlan M. Krumholz, Jure Leskovec, Eric J. Topol, et al.. Foundation models for generalist medical artificial intelligence. Nature, 2023. \url{https://doi.org/10.1038/s41586-023-05881-4}
\bibitem{ref2} Julián Acosta, Guido J. Falcone, Pranav Rajpurkar, Eric J. Topol. Multimodal biomedical AI. Nature Medicine, 2022. \url{https://doi.org/10.1038/s41591-022-01981-2}
\bibitem{ref3} Tao Tu, Shekoofeh Azizi, Danny Driess, Mike Schaekermann, Mohamed Amin, Pi-Chuan Chang, et al.. Towards Generalist Biomedical AI. NEJM AI, 2024. \url{https://doi.org/10.1056/aioa2300138}
\bibitem{ref4} Chunyuan Li, Cliff Wong, Sheng Zhang, Naoto Usuyama, Haotian Liu, Jianwei Yang, et al.. LLaVA-Med: Training a Large Language-and-Vision Assistant for Biomedicine in One Day. arXiv (Cornell University), 2023. \url{https://doi.org/10.48550/arxiv.2306.00890}
\bibitem{ref5} Michael Moor, Qian Huang, Shirley Wu, Michihiro Yasunaga, Cyril Zakka, Yash Dalmia, et al.. Med-Flamingo: a Multimodal Medical Few-shot Learner. arXiv (Cornell University), 2023. \url{https://doi.org/10.48550/arxiv.2307.15189}
\bibitem{ref6} Sheng Zhang, Yanbo Xu, Naoto Usuyama, Xu, Hanwen, Jaspreet Bagga, Robert Tinn, et al.. BiomedCLIP: a multimodal biomedical foundation model pretrained from fifteen million scientific image-text pairs. arXiv (Cornell University), 2023. \url{https://doi.org/10.48550/arxiv.2303.00915}
\bibitem{ref7} Benedikt Boecking, Naoto Usuyama, Shruthi Bannur, Daniel C. Castro, Anton Schwaighofer, Stephanie L. Hyland, et al.. Making the Most of Text Semantics to Improve Biomedical Vision-Language Processing. Lecture notes in computer science, 2022. \url{https://doi.org/10.1007/978-3-031-20059-5_1}
\bibitem{ref8} Yuhao Zhang, Hang Jiang, Yasuhide Miura, Christopher D. Manning, Curtis P. Langlotz. Contrastive Learning of Medical Visual Representations from Paired Images and Text. arXiv (Cornell University), 2020. \url{https://doi.org/10.48550/arxiv.2010.00747}
\bibitem{ref9} Shih-Cheng Huang, Liyue Shen, Matthew P. Lungren, Serena Yeung. GLoRIA: A Multimodal Global-Local Representation Learning Framework for Label-efficient Medical Image Recognition. 2021 IEEE/CVF International Conference on Computer Vision (ICCV), 2021. \url{https://doi.org/10.1109/iccv48922.2021.00391}
\bibitem{ref10} Zifeng Wang, Zhenbang Wu, D.C. Agarwal, Jimeng Sun. MedCLIP: Contrastive Learning from Unpaired Medical Images and Text. conference-paper, 2022. \url{https://doi.org/10.18653/v1/2022.emnlp-main.256}
\bibitem{ref11} Alec Radford, Jong Wook Kim, Chris Hallacy, Aditya Ramesh, Gabriel Goh, Sandhini Agarwal, et al.. Learning Transferable Visual Models From Natural Language Supervision. arXiv (Cornell University), 2021. \url{https://doi.org/10.48550/arxiv.2103.00020}
\bibitem{ref12} Alayrac, Jean-Baptiste, Jeff Donahue, Pauline Luc, Antoine Miech, Iain Barr, Yana Hasson, et al.. Flamingo: a Visual Language Model for Few-Shot Learning. arXiv (Cornell University), 2022. \url{https://doi.org/10.48550/arxiv.2204.14198}
\bibitem{ref13} Junnan Li, Dongxu Li, Silvio Savarese, Steven C. H. Hoi. BLIP-2: Bootstrapping Language-Image Pre-training with Frozen Image Encoders and Large Language Models. arXiv (Cornell University), 2023. \url{https://doi.org/10.48550/arxiv.2301.12597}
\bibitem{ref14} Haotian Liu, Chunyuan Li, Qingyang Wu, Yong Jae Lee. Visual Instruction Tuning. arXiv (Cornell University), 2023. \url{https://doi.org/10.48550/arxiv.2304.08485}
\bibitem{ref15} Ashish Vaswani, Noam Shazeer, Niki Parmar, Jakob Uszkoreit, Llion Jones, Aidan N. Gomez, et al.. Attention Is All You Need. Advances in Neural Information Processing Systems, 2017. \url{https://arxiv.org/abs/1706.03762}
\bibitem{ref16} Bjoern Menze, András Jakab, Stefan Bauer, Jayashree Kalpathy-Cramer, Keyvan Farahani, Justin Kirby, et al.. The Multimodal Brain Tumor Image Segmentation Benchmark (BRATS). IEEE Transactions on Medical Imaging, 2014. \url{https://doi.org/10.1109/tmi.2014.2377694}
\bibitem{ref17} Spyridon Bakas, Hamed Akbari, Aristeidis Sotiras, Michel Bilello, Martin Rozycki, Justin Kirby, et al.. Advancing The Cancer Genome Atlas glioma MRI collections with expert segmentation labels and radiomic features. Scientific Data, 2017. \url{https://doi.org/10.1038/sdata.2017.117}
\bibitem{ref18} Spyridon Bakas, Chiharu Sako, Hamed Akbari, Michel Bilello, Aristeidis Sotiras, Gaurav Shukla, et al.. The University of Pennsylvania glioblastoma (UPenn-GBM) cohort: advanced MRI, clinical, genomics, \& radiomics. Scientific Data, 2022. \url{https://doi.org/10.1038/s41597-022-01560-7}
\bibitem{ref19} Evan Calabrese, Javier Villanueva-Meyer, Jeffrey D. Rudie, Andreas M. Rauschecker, Ujjwal Baid, Spyridon Bakas, et al.. The University of California San Francisco Preoperative Diffuse Glioma MRI (UCSF-PDGM) Dataset. arXiv (Cornell University), 2021. \url{https://doi.org/10.48550/arxiv.2109.00356}
\bibitem{ref20} David N. Louis, Arie Perry, Pieter Wesseling, Daniel J. Brat, Ian A. Cree, Dominique Figarella-Branger, et al.. The 2021 WHO Classification of Tumors of the Central Nervous System: a summary. Neuro-Oncology, 2021. \url{https://doi.org/10.1093/neuonc/noab106}
\bibitem{ref21} Michael Weller, Martin J. van den Bent, Matthias Preusser, Émilie Le Rhun, Jörg C. Tonn, Giuseppe Minniti, et al.. EANO guidelines on the diagnosis and treatment of diffuse gliomas of adulthood. Nature Reviews Clinical Oncology, 2020. \url{https://doi.org/10.1038/s41571-020-00447-z}
\bibitem{ref22} Wenxuan Wang, Chen Chen, Meng Ding, Hong Yu, Sen Zha, Jiangyun Li. TransBTS: Multimodal Brain Tumor Segmentation Using Transformer. Lecture notes in computer science, 2021. \url{https://doi.org/10.1007/978-3-030-87193-2_11}
\bibitem{ref23} Ali Hatamizadeh, Vishwesh Nath, Yucheng Tang, Dong Yang, Holger R. Roth, Daguang Xu. Swin UNETR: Swin Transformers for Semantic Segmentation of Brain Tumors in MRI Images. Lecture notes in computer science, 2022. \url{https://doi.org/10.1007/978-3-031-08999-2_22}
\bibitem{ref24} Fabian Isensee, Paul F. Jaeger, Simon A. A. Kohl, Jens Petersen, Klaus H. Maier-Hein. nnU-Net: a self-configuring method for deep learning-based biomedical image segmentation. Nature Methods, 2020. \url{https://doi.org/10.1038/s41592-020-01008-z}
\bibitem{ref25} Jun Ma, Yuting He, Feifei Li, Lin Han, Chenyu You, Bo Wang. Segment anything in medical images. Nature Communications, 2024. \url{https://doi.org/10.1038/s41467-024-44824-z}
\bibitem{ref26} Chuan Guo, Geoff Pleiss, Yu Sun, Kilian Q. Weinberger. On Calibration of Modern Neural Networks. arXiv (Cornell University), 2017. \url{https://doi.org/10.48550/arxiv.1706.04599}
\bibitem{ref27} Sebastian Farquhar, Jannik Kossen, Lorenz Kuhn, Yarin Gal. Detecting hallucinations in large language models using semantic entropy. Nature, 2024. \url{https://doi.org/10.1038/s41586-024-07421-0}
\bibitem{ref28} Matteo Fontana, Gianluca Zeni, Simone Vantini. Conformal prediction: A unified review of theory and new challenges. Bernoulli, 2022. \url{https://doi.org/10.3150/21-bej1447}
\bibitem{ref29} John Mongan, Linda Moy, Charles E. Kahn. Checklist for Artificial Intelligence in Medical Imaging (CLAIM): A Guide for Authors and Reviewers. Radiology Artificial Intelligence, 2020. \url{https://doi.org/10.1148/ryai.2020200029}
\bibitem{ref30} Professor Gary S. Collins, Karel G.M. Moons, Paula Dhiman, Richard D Riley, Andrew L. Beam, Ben Van Calster, et al.. TRIPOD+AI statement: updated guidance for reporting clinical prediction models that use regression or machine learning methods. BMJ, 2024. \url{https://doi.org/10.1136/bmj-2023-078378}
\bibitem{ref31} Richard J. Chen, Judy J. Wang, Drew F. K. Williamson, Tiffany Chen, Jana Lipková, Ming Y. Lu, et al.. Algorithmic fairness in artificial intelligence for medicine and healthcare. Nature Biomedical Engineering, 2023. \url{https://doi.org/10.1038/s41551-023-01056-8}
\bibitem{ref32} Paul Hager, Friederike Jungmann, Robbie Holland, Kunal Bhagat, Inga Hubrecht, Manuel Knauer, et al.. Evaluation and mitigation of the limitations of large language models in clinical decision-making. Nature Medicine, 2024. \url{https://doi.org/10.1038/s41591-024-03097-1}
\bibitem{ref33} Patrick Lewis, Ethan Perez, Aleksandra Piktus, Fabio Petroni, Vladimir Karpukhin, Naman Goyal, et al.. Retrieval-Augmented Generation for Knowledge-Intensive NLP Tasks. UCL Discovery (University College London), 2020. \url{https://discovery.ucl.ac.uk/id/eprint/10100504/}
\bibitem{ref34} Zhihong Chen, Yan Song, Tsung-Hui Chang, Xiang Wan. Generating Radiology Reports via Memory-driven Transformer. conference-paper, 2020. \url{https://doi.org/10.18653/v1/2020.emnlp-main.112}
\bibitem{ref35} Alistair E. W. Johnson, Tom Pollard, Seth J. Berkowitz, Nathaniel R. Greenbaum, Matthew P. Lungren, Chih-Ying Deng, et al.. MIMIC-CXR: A large publicly available database of labeled chest radiographs. arXiv (Cornell University), 2019. \url{https://arxiv.org/pdf/1901.07042.pdf}
\bibitem{ref36} Saahil Jain, Ashwin Agrawal, Adriel Saporta, Steven Qh Truong, Du Nguyen Duong, Tan Bui, et al.. RadGraph: Extracting Clinical Entities and Relations from Radiology Reports. arXiv (Cornell University), 2021. \url{https://doi.org/10.48550/arxiv.2106.14463}
\bibitem{ref37} Akshay Smit, Saahil Jain, Pranav Rajpurkar, Anuj Pareek, Andrew Y. Ng, Matthew P. Lungren. CheXbert: Combining Automatic Labelers and Expert Annotations for Accurate Radiology Report Labeling Using BERT. arXiv (Cornell University), 2020. \url{https://doi.org/10.48550/arxiv.2004.09167}
\end{thebibliography}
\end{document}